\pdfoutput=1
\documentclass[11pt]{article}

\usepackage[preprint]{acl}  % review, final, preprint

\usepackage[utf8]{inputenc}
\usepackage[T1]{fontenc}
\usepackage{times}
\usepackage{latexsym}
\usepackage{microtype}
\usepackage{inconsolata}
\usepackage{graphicx}

\usepackage{hyperref}
\usepackage{url}
\usepackage{booktabs}
\usepackage{amsbsy}
\usepackage{amsfonts}
\usepackage{amsmath}
\usepackage{amssymb}
\usepackage{centernot}
\usepackage{mathtools}
\usepackage{dsfont}
\usepackage{csquotes}
\usepackage{adjustbox}
\usepackage{array}
\usepackage{multirow}
\usepackage{caption}
\usepackage{subcaption}
\usepackage{float}
\usepackage{enumitem}
\usepackage{xcolor}
\usepackage{soul}
\usepackage{nicefrac}
\usepackage{arydshln}
\usepackage{stfloats}

\definecolor{darkterracotta}{rgb}{0.8, 0.31, 0.36}
\definecolor{deepcarmine}{rgb}{0.66, 0.13, 0.24}
\definecolor{iris}{rgb}{0.35, 0.31, 0.81}
\definecolor{pinegreen}{rgb}{0.0, 0.47, 0.44}
\definecolor{lightgray}{gray}{0.55}
\hypersetup{
    colorlinks=true,
    citecolor=iris,
    linkcolor=darkterracotta,
    filecolor=darkterracotta,
    urlcolor=pinegreen,
    pdftitle={The Dialect Tax},
    pdfpagemode=FullScreen,
}

\newcommand{\norm}[1]{\left\lVert#1\right\rVert}

\title{The Dialect Tax:\\Dialectal Biases Persist throughout the Language Modeling Pipeline}

\author{
    \textbf{Elle}\,\textsuperscript{\includegraphics[height=\baselineskip]{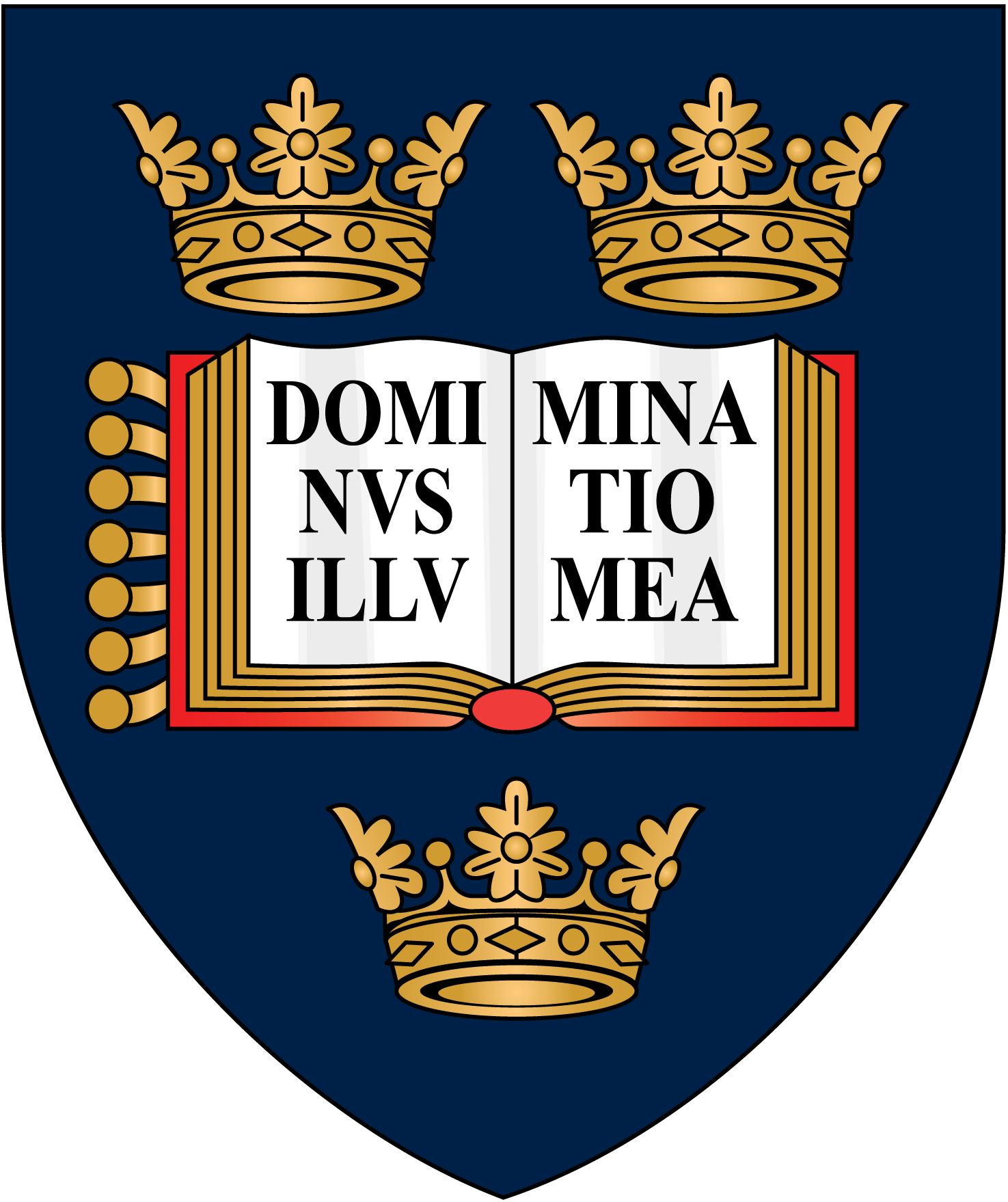}}\\
    \textsuperscript{\includegraphics[height=\baselineskip]{figures/oxford.png}}\,University of Oxford, Department of Computer Science\\
    \hypersetup{hidelinks}\small{\href{mailto:elle.yang@cs.ox.ac.uk}{\texttt{elle.yang@cs.ox.ac.uk}}}
}

\begin{document}
\maketitle
\begin{abstract}
Systematic dialectal performance gaps in language models (LMs) are well documented, but the source of these disparities within the modern language modeling pipeline remains unclear. Our study traces this \enquote{dialect tax} across the natural language processing pipeline. Using parallel English dialect corpora that hold meaning fixed while varying surface form, we first confirm that LMs recognize matched Standard American English (SAE) and dialectal texts as semantically equivalent. However, we discover further representational gaps corresponding to downstream performance gaps. Across model families and generations, modern LMs still encode dialectal texts unequally during tokenization, pre-training, post-training, and inference. Strikingly, bypassing traditional subword segmentation via a character-level counterfactual tokenizer removes neither input and output asymmetries nor dialectal accuracy gaps. During pre-training, dialect pairs induce more divergent gradient updates than pairs of entirely unrelated SAE documents, indicating that models find semantically equivalent dialectal content harder to learn from than unrelated SAE documents. During post-training, reward models show contextual, unstable dialect preferences, assigning higher values to isolated AAVE-exclusive tokens than to SAE-exclusive tokens, while full reasoning contexts receive task- and model-dependent dialect penalties. Overall, our findings suggest that the dialect tax is encoded and accumulated not by any one step in isolation, but at every step of the language modeling process.
\end{abstract}

% TL;DR: We trace dialectal biases through every stage of the language modeling pipeline -- tokenization, pre-training, post-training, and inference -- and find that each stage encodes the bias while none corrects it.

\section{Introduction}

AI systems, particularly language models (LMs), have become technological staples worldwide. Although advances in deep learning for natural language processing (NLP) have improved the performance of these systems, these technologies continue to exhibit societal biases at every stage of the language modeling process \citep{okpala2022aaebert,buyl2024llmsreflectcreatorideology,joshi2024naturallanguageprocessingdialects}.

A prominent form of linguistic variation is dialectal variation, characterized by macro-social factors such as geography, socioeconomic class, ethnicity, gender, and age \citep{haugen1966dialect,haugh2012impolitenessacrossdialects}. The resulting dialects are primarily separated into two categories: standard and non-standard \citep{trudgill2004dialects}. By definition, \enquote{standard} dialects spoken by the majority receive preferential treatment from governmental and educational institutions, providing speakers of standard dialects with more socioeconomic opportunities than speakers of minority dialects \citep{trudgill1979policies}. This creates a circular compounding effect that systemically disadvantages minority groups, leaving historically marginalized communities that speak \enquote{non-standard} language varieties vulnerable to further discrimination by modern AI systems trained on historical data \citep{blodgett2016socialmedia,jurgens2017variability,kantharuban2023dialectgap,drozdzowicz2024linguisticdiscrimination,nayeem2026englishllmsprefertriangulating}.

Previous research has shown that components across the language modeling stack behave suboptimally on uncommon linguistic forms -- such as low-resource languages and dialects -- resulting in problematic biases and disparities in cost, latency, and quality that exacerbate economic divides in the accessibility of language technologies. These biases have been identified in tokenizers \citep{ahia2023languagescostsametokenization,petrov2023languagemodeltokenizersintroduce}, LMs \citep{redial}, and reward models \citep[RMs;][]{mire2025rejecteddialects}. Systematic discrepancies in token representations of raw text across different social groups are commonly assumed to cause downstream performance issues, yet whether these biases originate from the tokenizer remains an open question.

In standard NLP pipelines, tokenization is the first step in transforming human-readable text into a machine-readable format \citep{grefenstette1999}. Modern LMs use subword tokenization, which decomposes text into a sequence of subword units, or \enquote{subtokens}, drawn from a tokenizer vocabulary. Crucially, while the exact algorithmic details differ, tokenizers work by optimizing a segmentation model on a training corpus, a process that does not explicitly account for uncommon linguistic forms. For example, \enquote{buildin'} might be tokenized as \texttt{[`build', `in', ``'{}'']}, whereas \enquote{building} might be represented by the single token \texttt{[`building']}. This algorithmic approach is a double-edged sword that can either improve model performance by promoting the recognition of form-based patterns \citep{wegmann2025languagevariation} or reduce model performance by producing unwanted artifacts \citep{hofmann2022embarrassinglysimple,arnett2024languagemodelsperformworse,schmidt2024tokenizationcompression}.

To advance toward fairer language technologies that minimize the undesirable effects of intra-language variation, we aim to identify the sources of dialectal bias. This work uses parallel dialect corpora in English, the highest-resource language, to trace the \enquote{dialect tax} through every stage of the modern NLP pipeline: tokenization, pre-training, post-training, and inference. By comparing English to English, our analysis removes the confounding factor of byte premiums \citep{arnett2024bitproblemmeasurementdisparities}.

We begin by establishing that models recognize matched SAE and dialectal texts as semantic equivalents to a greater extent than they recognize translations or character-perturbed copies of the same text (\S\ref{same_meaning_different_taxes}). Nonetheless, our experiments document the persistence of dialectal biases across the latest generation of tokenizers and reproduce reasoning gaps across currently popular LMs (\S\ref{dialectal_biases}). Our results verify that, compared with Standard American English (SAE), dialects such as African American Vernacular English (AAVE) -- which are often already stigmatized -- continue to suffer from suboptimal model performance.

To investigate the origins of dialect biases, we first isolate the tokenizer's causal role using a character-level tokenization counterfactual (\S\ref{token_isolation}). By removing subword segmentation at inference time, we narrow the input-side dialect gap. This character-level tokenization intervention, however, leaves accuracy and output behavior largely unchanged. These results provide insight into how dialect biases persist beyond the tokenizer into the model's learned weights, necessitating further scrutiny of training dynamics. Indeed, we present empirical evidence that dialect taxes are introduced during both pre-training and post-training (\S\ref{dialect_taxes}). Across dialect pairs, we observe divergence in base-model gradients, LM hidden states, and RM preferences. Paired SAE and AAVE inputs yield gradient updates that are more dissimilar than those delivered by two entirely unrelated SAE documents, suggesting that dialectal texts consistently carry a higher prediction cost regardless of problem difficulty. RMs reward dialect-exclusive tokens in isolation yet penalize them in context, which risks propagating these biases into deployed LMs. Collectively, we show that dialectal biases originate not from tokenizers alone but accumulate across LMs and RMs. Addressing the dialect tax therefore requires new recipes for model training.

\section{Same Meaning, Higher Taxes}
\label{same_meaning_different_taxes}

Before tracing the source of dialect biases, we reduce concerns that observed gaps are explained by gross semantic mismatch. We verify that a text-embedding model assigns high semantic similarity to matched dialect pairs, meaning dialect-dependent behavior reflects surface-form biases.

\begin{figure}[ht]
\centering
\includegraphics[width=\linewidth]{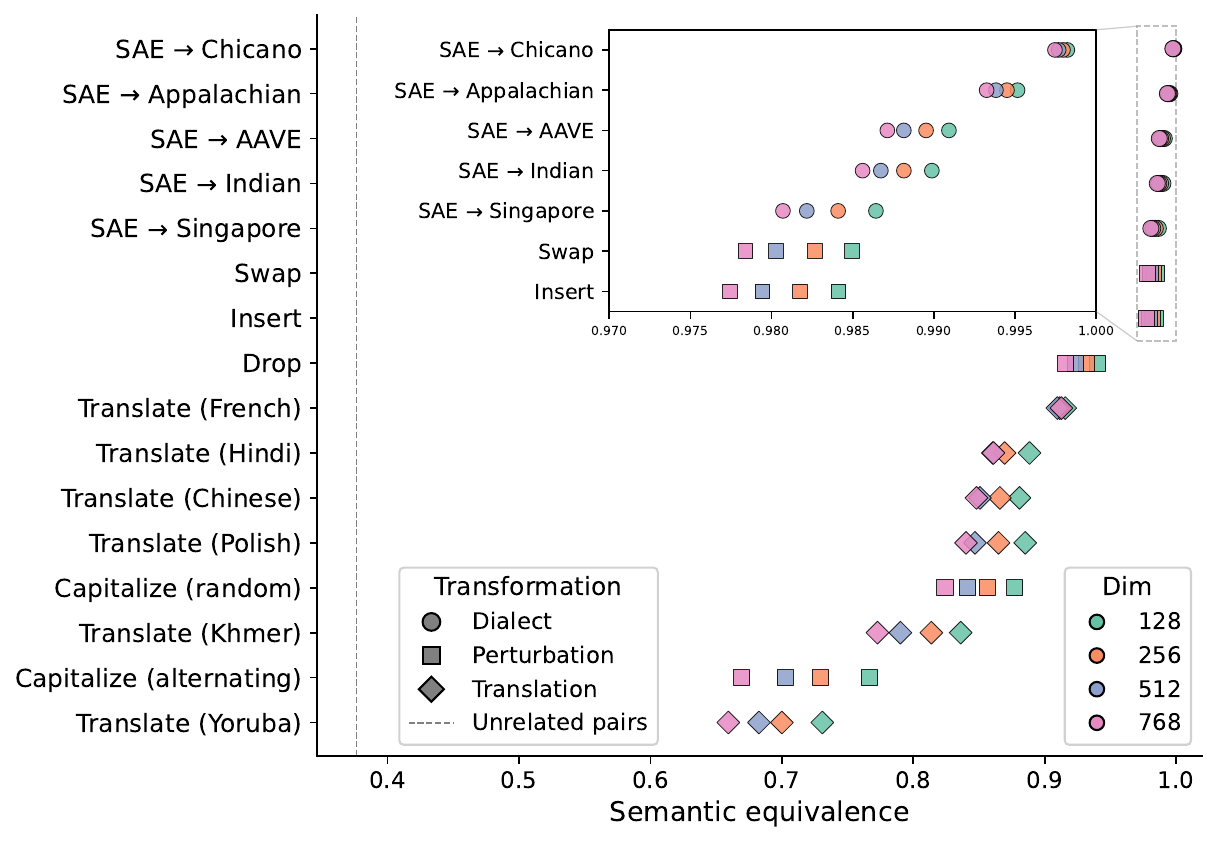}
\caption{\textbf{Models understand semantic equivalence yet penalize surface form.} We visualize the semantic equivalence of various text transformations on \textsc{MultiVALUE}. All dialect pairs achieve high similarities exceeding every perturbation and translation baseline.}
\label{figure:semantic_equivalence_surface_difference}
\end{figure}

\vspace{.5\baselineskip}

\noindent\textbf{Datasets.} \textsc{ParallelAAVE} \citep{parallelaave} and \textsc{MultiVALUE} \citep{multivalue} offer parallel texts between SAE and American dialects (Table~\ref{table:data-parallelaave-multivalue}). Each \textsc{ParallelAAVE} pair contains a tweet written in AAVE and its corresponding human rewrite in SAE. Each \textsc{MultiVALUE} pair uses a rule-based translation system to synthetically generate SAE texts from \textsc{CoQA} \citep{reddy2018coqa} into AAVE, Appalachian, Chicano, Indian, and Singaporean dialects.

\vspace{.5\baselineskip}

\noindent\textbf{Setup.} We use \textsc{EmbeddingGemma} \citep{embeddinggemma}, a 300M-parameter text embedding model based on the \textsc{Gemma-3} LM family, which produces numerical vector representations of four nested dimensions ($768$, $512$, $256$, $128$) via Matryoshka Representation Learning \citep{kusupati2024matryoshkarepresentationlearning}.\footnote{\textsc{EmbeddingGemma} was chosen to measure similarity, as it is explicitly optimized for tasks such as semantic similarity and is derived from one of the LM families used the analyses of later sections.} We quantify semantic equivalence between two texts by calculating the cosine similarity $\text{sim}(x, y) = \langle x, y \rangle / (\norm{x}_2 \norm{y}_2)$ between their embeddings $x$ and $y$. We compare every SAE sample in \textsc{MultiValue} and \textsc{ParallelAAVE} to three types of text transformations, which serve as a calibration baseline for changing surface form while preserving meaning to assess dialectal treatment.

\vspace{.5\baselineskip}

\noindent\textbf{Transformations.} We identify a set of transformations $\mathcal{T} = \{\tau_1, \tau_2, \ldots, \tau_m\}$ that preserve semantic content while perturbing surface form. For each text sample $x_i$ and transformation $\tau_j$, we compute $\text{sim}(x_i, \tau_j(x_i))$. (1) \textit{Dialects.} We use parallel texts. (2) \textit{Perturbations.} We apply character-level perturbations: swaps ($\mathbb{P} = 0.05$), drops ($\mathbb{P} = 0.15$), inserts ($\mathbb{P} = 0.05$), and capitalizations (alternating or $\mathbb{P} = 0.5$). (3) \textit{Translations.} We translate the original SAE text into six languages with high (French, Chinese), medium (Hindi, Polish), or low (Khmer, Yoruba) resources.

\vspace{.5\baselineskip}

\noindent\textbf{Dialects show semantic equivalence.} We find evidence of semantic invariance in embedding models under surface-form transformations. Figure~\ref{figure:semantic_equivalence_surface_difference} compares per-sentence cosine similarities between SAE and dialect rewrites with those of SAE and perturbation or translation controls. On \textsc{MultiVALUE}, all five dialect pairs achieve similarities above $0.98$, significantly exceeding every perturbation and translation control baseline exhibiting similarities between $0.659$ to $0.978$. We confirm statistical significance (Holm-adjusted $p < 0.001$) using a one-sided paired Wilcoxon signed-rank test over $429$ aligned sentences to confirm all five dialect conditions were more similar to SAE than each of the controls. On \textsc{ParallelAAVE}, the SAE to AAVE similarity is $0.92$, which significantly (Holm-adjusted $p < 0.001$) exceeds character deletion ($0.840$), capitalization ($0.702$ to $0.786$), and translations ($0.534$ to $0.872$), but is lower than character swapping ($0.956$) and insertion ($0.951$). All values are far above the null baseline of unrelated document pairs ($\mu = 0.41$). This floor establishes that dialect transformations, along with their original SAE texts, have scores that signal genuine similarities between each pair.

We check for the confound that the embedding model assigns higher similarities for texts that are more predictable to the model rather than for texts that preserve meaning. If the confound were true, transformations that are more surprising (higher perplexity) to a model would predict lower similarity. Instead, similarity tracks semantic content rather than LM surprisal (Table~\ref{table:perturbation_perplexity_ratios}). Across all transformations, mean similarity correlates positively -- rather than negatively -- with log median per-token input-perplexity ratio (\textsc{MultiVALUE}: $\rho = +0.54$, \textsc{ParallelAAVE}: $\rho = +0.43$).

\section{Models (Still) Have Dialectal Biases}
\label{dialectal_biases}

\begin{figure*}[ht]
\centering
\begin{subfigure}{0.49\textwidth}
    \centering
    \includegraphics[width=\textwidth]{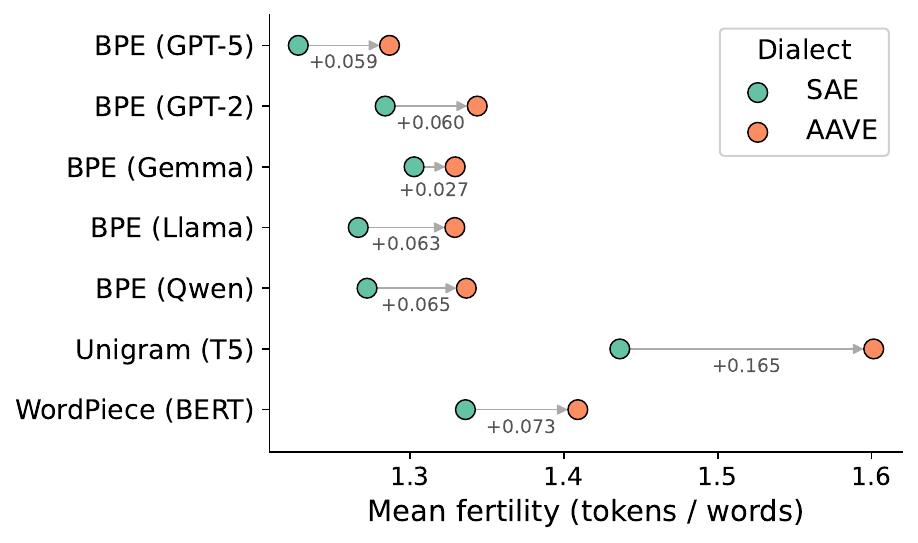}
    \caption{Mean fertility of \textsc{ParallelAAVE}}
    \label{figure:token_bias_disparities-parallelaave_fertility}
\end{subfigure}
\hfill
\begin{subfigure}{0.49\textwidth}
    \centering
    \includegraphics[width=\textwidth]{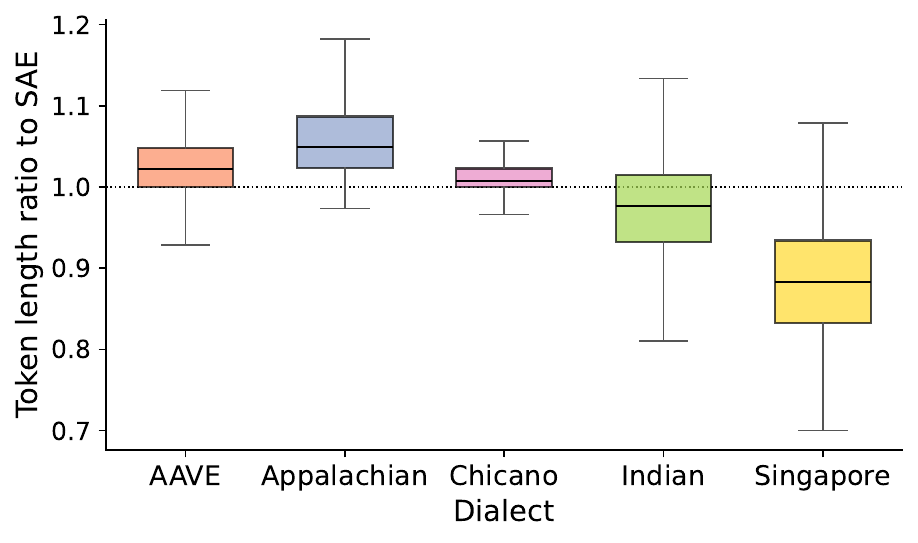}
    \caption{Token length ratio of \textsc{MultiVALUE}}
    \label{figure:token_bias_disparities-multivalue_token_ratio}
\end{subfigure}
\caption{\textbf{Modern tokenizers consistently exhibit dialectal biases.} (\subref{figure:token_bias_disparities-parallelaave_fertility}) The mean fertility of the three tokenization algorithms on \textsc{ParallelAAVE} shows a statistically significant difference between SAE and AAVE dialects. (\subref{figure:token_bias_disparities-multivalue_token_ratio}) The ratio of \textsc{BPE} token lengths on various dialects to that of SAE on \textsc{MultiVALUE} reveals a consistent dialectal tokenization performance gap that loosely parallels current income gaps of minority groups within the US (\S\ref{appendix:tokenization-biases}).}
\label{figure:token_bias_disparities}
\end{figure*}

While models have improved and hill-climbed various benchmarks, we verify that newer generations of models still exhibit token biases on information compression and downstream reasoning tasks.

\subsection{Tokenizers Have Dialectal Biases}

\noindent\textbf{Tokenizers.} We explore the three most popular tokenization methods: byte-pair encoding (\textsc{BPE}) \citep{sennrich2015bpe}, \textsc{Unigram} \citep{kudo2018unigram}, and \textsc{WordPiece} \citep{schuster2012wordpiece}. We evaluate tokenization metrics using the tokenizers of popular LMs (Table~\ref{table:tokenizers}). \textsc{BPE} variants include \textsc{GPT-5}, \textsc{GPT-2}, \textsc{Gemma-3}, \textsc{Llama-3}, and \textsc{Qwen-3} models. \textsc{Unigram} uses the \textsc{T5} model. \textsc{WordPiece} uses the \textsc{BERT} model. We drop the LM family number where the value is obvious.

\vspace{.5\baselineskip}

\noindent\textbf{Metrics.} We measure token biases through eight metrics (Table~\ref{table:tokenization-bias-metrics}) and showcase the token length and fertility (number of tokens divided by the number of words) in the main section. We assess tokenizer fairness by evaluating whether these metrics achieve parity on parallel texts across all dialects, i.e. metrics should be comparable when conditioned on the same content.

\vspace{.5\baselineskip}

\noindent\textbf{Results.} Tokenizers consistently produce unequal representations across English dialects. Figure~\ref{figure:token_bias_disparities-parallelaave_fertility} demonstrates that dialectal tokenization biases persist across modern tokenizer and model families. Across all tokenizers, AAVE texts have higher mean fertility ($\downarrow$ better) than their SAE counterparts, with an average gap of $+0.07$ tokens per word. Figure~\ref{figure:token_bias_disparities-multivalue_token_ratio} suggests that this tokenization bias could systematically amplify existing sociodemographic bias. We analyze the five non-SAE dialects in \textsc{MultiVALUE} by computing the per-sample ratio of dialect token length to the paired SAE token length, which showcases a consistent tokenization bias ranking: Appalachian, AAVE, and Chicano English texts incur increased token lengths (medians above parity), while Indian and Singaporean English texts have more compact token lengths (medians below parity) than SAE. This ranking is shown in Figure~\ref{figure:tokenizer-dialect-bias-ratio-multivalue} to be stable across all tokenizer families, indicating that the bias is structural to the dialectal text rather than an artifact of any particular algorithm. Notably, this tokenization performance gap loosely parallels the income distribution of minority groups within the US (Figure~\ref{figure:tokenization-bias-and-income}).

\subsection{LMs Have Dialectal Biases}

\noindent\textbf{Setup.} We modify the \textsc{ReDial} \citep{redial} dataset, a reasoning benchmark containing parallel query pairs in SAE and AAVE, and reproduce their original results using a newer set of models. Changes include reformatting, instruction rewording, and correctness validation, with additional details left to Appendix~\ref{appendix:redial}. We evaluate three open-source LM families (\textsc{Llama-3}: 8B, 70B; \textsc{Gemma-3}: 12B, 27B; \textsc{Qwen-3}: 8B, 27B) and OpenAI's \textsc{GPT} family (\textsc{GPT-5.5}, \textsc{GPT-5.5 Mini}) to benchmark both SAE and AAVE dialects ($\sim300$ samples per condition) on four tasks (math, algorithm, logic, planning) using chain-of-thought (\enquote{CoT}) or no CoT (\enquote{na\"ive}) reasoning.

\vspace{.5\baselineskip}

\noindent\textbf{Results.} LMs continue to show weaker reasoning capabilities on AAVE text than SAE text, as shown in Table~\ref{table:benchmark-redial}. The SAE-AAVE reasoning gap exists for all model families and sizes, indicating the dialect tax remains an unsolved issue.

\section{Isolating the Tokenizer from the Model}
\label{token_isolation}

\begin{figure*}[b]
\centering
\includegraphics[width=\textwidth]{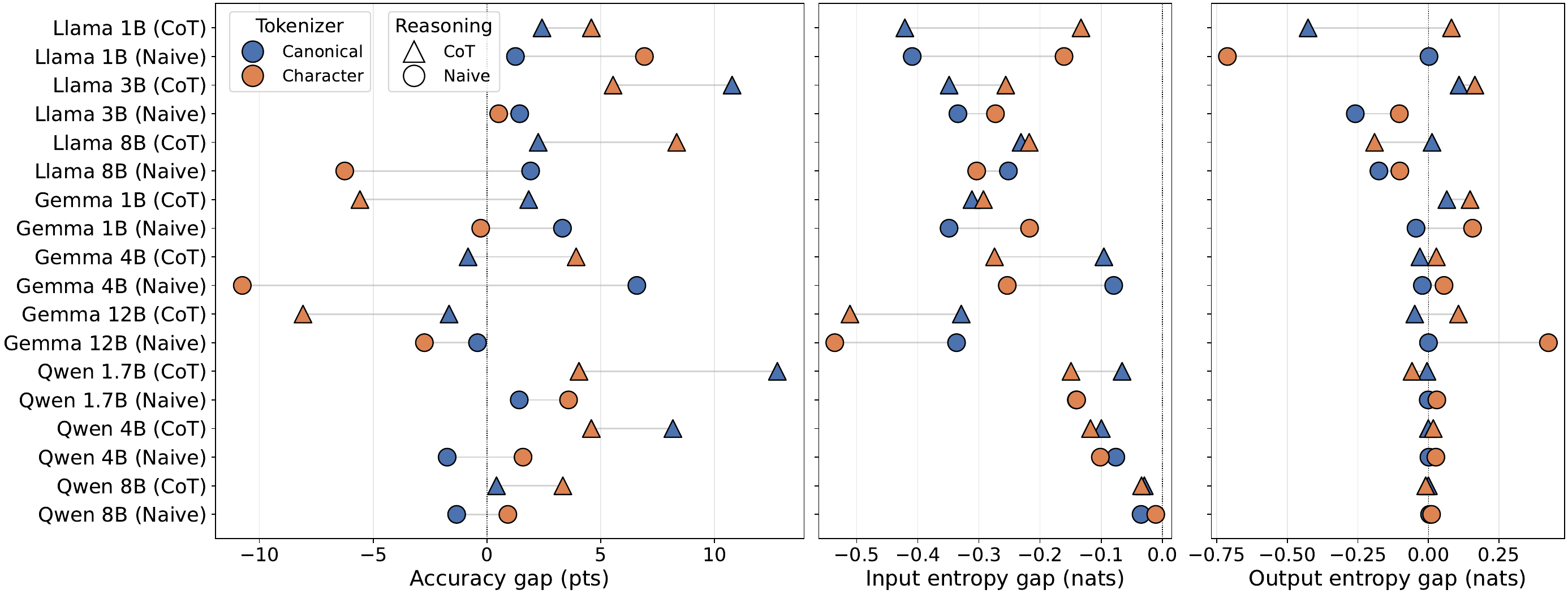}
\caption{\textbf{Character tokenization does not systematically narrow dialect gaps.} We plot per-model SAE~$-$~AAVE effect sizes under canonical (blue) and character-level (orange) tokenization, split by reasoning strategy ($\triangle = \text{CoT}$, $\bigcirc = \text{na\"ive}$). Grey lines connect paired points for each model. \textit{Left:} The accuracy gap does not systematically change under character tokenization. \textit{Center:} The input entropy gap $\Delta_H$ changes inconsistently, confirming that character tokenization fails to equalize how the model processes dialect inputs. \textit{Right:} The output entropy gap $\Delta_H$ persists under both tokenizations, suggesting generation-time dialect bias is independent of the tokenizer.}
\label{figure:character_accuracy_entropy_gaps}
\end{figure*}

We now ask if the dialect tax is generated by subword segmentation at inference or encoded in a model's priors from training. We disentangle a \textit{tokenizer-induced} from a \textit{model-induced} account by forcing LMs to evaluate the same prompts under character-level tokenization that bypasses subword merges entirely. Gaps that vanish implicate the tokenizer, while gaps that persist implicate the LM.

\vspace{.5\baselineskip}

\noindent\textbf{Setup.} We employ either a canonical tokenizer or a character tokenizer at inference time (\S\ref{appendix:character-tokenization}) on nine instruction-tuned LMs across three model families (\textsc{Llama-3}: 1B, 3B, 8B; \textsc{Gemma-3}: 1B, 4B, 12B; \textsc{Qwen-3}: 1.7B, 4B, 8B). Our method indexes off the results by \citet{zheng2026brokentokenslanguagemodel} showing that swapping a model's canonical tokenizer retains the majority of changes in that model's behavior, and non-canonical tokenizations can outperform canonical tokenizations in some domains that require non-canonical representations of text (e.g., counting characters, arithmetic). Thus, given the known disparities between the tokenizations of dialectal text and previous findings that suboptimal tokenization leads to poor performance, character tokenization helps us isolate LM performance from tokenizer effects on SAE versus dialectal texts.

Using the \textsc{ReDial} dataset, we measure: accuracy, input entropy, and output entropy. As LM behavior is necessarily entangled with its canonical tokenizer, we are primarily concerned with the relative -- rather than the absolute -- change in the paired dialect gap. The per-token entropy gap is denoted by $\Delta_H = \mathbb{E}[H(p_\theta(\cdot | x_{<t}))]_{\text{SAE}} - \mathbb{E}[H(p_\theta(\cdot | x_{<t}))]_{\text{AAVE}}$, where $H(\cdot)$ is Shannon entropy over the predictive distribution. If the dialect tax were purely tokenizer-induced, character tokenization should close the gap on the three metrics.

\vspace{.5\baselineskip}

\begin{table*}[ht]
\centering
\resizebox{.7\linewidth}{!}{
\begin{tabular}{lcccccccccc}
\toprule
& \multicolumn{3}{c}{\textsc{Llama}} & \multicolumn{3}{c}{\textsc{Gemma}} & \multicolumn{3}{c}{\textsc{Qwen}} \\
\cmidrule(lr){2-4} \cmidrule(lr){5-7} \cmidrule(lr){8-10}
& 1B & 3B & 8B & 1B & 4B & 12B & 1.7B & 4B & 8B \\
\midrule
Canonical (\%) & 99.0 & 99.9 & 99.4 & 99.8 & 99.8 & 100.0 & 99.9 & 100.0 & 100.0 \\
Character (\%) & 94.5 & 98.3 & 99.6 & 86.4 & 99.5 & 99.5 & 98.5 & 97.3 & 97.5 \\
$\Delta$ & $-4.5$ & $-1.6$ & $+0.2$ & $-13.4$ & $-0.3$ & $-0.5$ & $-1.3$ & $-2.7$ & $-2.5$ \\
\bottomrule
\end{tabular}
}
\caption{\textbf{Dialects are linearly decodable from hidden states under both tokenizations.} We list the five-fold cross-validation accuracy of logistic regression predicting dialect from the answer-step hidden state. Character tokenization slightly reduces separability, but all models remain far above chance (50\%).}
\label{tab:hidden_separability}
\end{table*}

\noindent\textbf{Behavioral disparities remain.} Dialect gaps survive forced character tokenization. These results reveal that canonical subword segmentation at inference time does not fully explain the dialectal disparity, suggesting that the dialect tax is at least partly model-induced. None of the aforementioned metrics shrink systematically under a character-level tokenizer, as pictured in Figure~\ref{figure:character_accuracy_entropy_gaps}. The input entropy gap and its directionality persist under character tokenization, with AAVE positions carrying higher mean per-token input entropy than its paired SAE counterpart in $93.0\%$ of paired samples under both tokenizations (one-sided Wilcoxon signed-rank tests, both $p < 0.001$), with $92.3\%$ sign agreement across $19{,}995$ pairs. The relative magnitude differences between both input entropy and output entropy of dialectal texts are mixed, suggesting that subword segmentation is not the sole contributor of the dialect tax.

Crucially, model downstream behavior is unaffected. The accuracy gap does not systematically change under character tokenization, as across nine models, five widen and four narrow, yielding statistically insignificant sign and paired $t$-test on gap magnitude. Our results suggest that subword segmentation inflates the gap but does not fully account for it. The output entropy gap at the answer-extraction step $H(p_\theta(\cdot | x, y_{<t^*}))$ is unchanged by the tokenizer swap across all model families. Character tokenization fails to narrow the input gap, output gap, nor accuracy disparities. This dissociation holds for individual SAE-AAVE texts, where input and output $\Delta_H$ changes are uncorrelated across $n = 13{,}197$ paired items (Pearson $r = +0.001$, $p = 0.91$). The persistence of these output gaps after bypassing subword segmentation suggests that they are attributable to the model's learned parameters rather than the tokenizer at inference time.

\vspace{.5\baselineskip}

\noindent\textbf{Representational differences remain.} Our experiments show that an LM's internal representations preserve dialect identity under character tokenization. The model-induced account predicts that the dialect should be recoverable from internal representations even after removing the tokenizer's contribution, which we test by training a logistic regression classifier to predict the dialect $d \in \{\text{AAVE}, \text{SAE}\}$ from the answer-step hidden state $\mathbf{h}_{t^*} \in \mathbb{R}^{d_\text{model}}$ under five-fold cross-validation. Table~\ref{tab:hidden_separability} compares the results from the classifier. Under canonical tokenization, classification accuracy is above $99\%$ across all nine models. Under character tokenization, it remains above $86\%$. Even the largest drop, \textsc{Gemma}~$1$B with $-13.4$ percentage points (pp), stays far above the $50\%$ chance baseline. These results are stable across model family, scale, and reasoning strategy, supporting the claim that a model's internal representations encode dialect identity within the learned features of $\mathbf{h}_{t^*}$ independently of the subword boundaries used to produce it. Since removing canonical subword segmentation reduces but does not eliminate dialect separability, we conclude that the dialect tax is not confined to the canonical tokenizer but extends into downstream model representations.

\section{Dialect Taxes Exist Despite Training}
\label{dialect_taxes}

We build on the evidence that learned representations exhibit dialect taxes by exposing the same disparities in pre-training gradients (\S\ref{dialect_taxes_pretraining}), post-training rewards(\S\ref{dialect_taxes_posttraining}), and inference-time hidden states and output distributions (\S\ref{dialect_taxes_inference}).

\subsection{Pre-Training Gradients}
\label{dialect_taxes_pretraining}

Similar content might be expected to induce similar learning signals during training. In other words, two inputs that pose the same task and have the same target answer, differing only by a meaning-preserving surface rewrite, carry the same ``knowledge'' to be learned. Thus, a model that represents a task in a dialect-invariant way should produce nearly identical gradient for both; what the model learns from a document ought not to depend on the dialect in which the text is written. Under dialect-invariant learning, matched documents in different dialects should yield greater gradient alignment than unrelated documents in the same dialect. We hypothesize that matched semantic content should provide an alignment advantage after controlling for dialect. We test this hypothesis for semantically equivalent dialect pairs by measuring their per-document gradient signatures.

\vspace{.5\baselineskip}

\noindent\textbf{Setup.} In accordance with previous sections, we compute the gradient projections for nine base models across three LM families (\textsc{Llama-3}: 1B, 3B, 8B; \textsc{Gemma-3}: 1B, 4B, 12B; \textsc{Qwen-3}: 1.7B, 4B, 8B). We use the modified \textsc{ReDial} dataset (\S\ref{appendix:redial}) with randomized multiple-choice answers.

\vspace{.5\baselineskip}

\noindent\textbf{Metrics.} Following \citet{kazdan2026scaledependentdataduplication}, we represent each document by the full-parameter gradient of its causal LM loss. For input $x$, this loss is the average next-token cross-entropy:
\begin{equation}
    \ell(x; \theta) = \frac{1}{|x|} \sum_{u=1}^{|x|} \text{CE}(f_\theta(x)_u, x_{u+1}).
\end{equation}
The document gradient is then:
\begin{equation}
    g(x; \theta) = \nabla_\theta \ell(x; \theta).
\end{equation}

Storing full gradients is impractical for billion-parameter models, so we compress each gradient via \textsc{CountSketch} \citep{spring2019countsketch}, a randomized linear projection that preserves inner products in expectation. Each parameter gradient coordinate is assigned element-wise to one of $d = 8{,}192$ buckets, multiplied by a random sign, and summed with other coordinates in that bucket. This yields a sketch $\hat{g}(x; \theta) \in \mathbb{R}^d$ providing a compact approximation of the original gradient space.

We measure the cosine similarity between paired SAE-AAVE gradient projections for each sample $i$, $s_i^+ = \text{sim}(\hat{g}(x_i^{\text{SAE}}; \theta), \, \hat{g}(x_i^{\text{AAVE}}; \theta))$.

To calibrate these similarities, we establish a null baseline mean $\mu^-$ and standard deviation $\sigma^-$ from the cosine similarity of gradient projections between unrelated SAE documents. We use this baseline to compute $z$-scores, $z = (\mu^+ - \mu^-) / \sigma^-$, where $\mu^+$ is the mean paired similarity. A $z$-score significantly above zero indicates that dialect pairs are more similar to each other than to unrelated texts, that is, the model recognizes shared content despite dialectal variation. A $z$-score near zero indicates the dialect shift is as disruptive to the gradient as switching to an entirely different text.

\vspace{.5\baselineskip}

\begin{figure}[t]
\centering
\includegraphics[width=\linewidth]{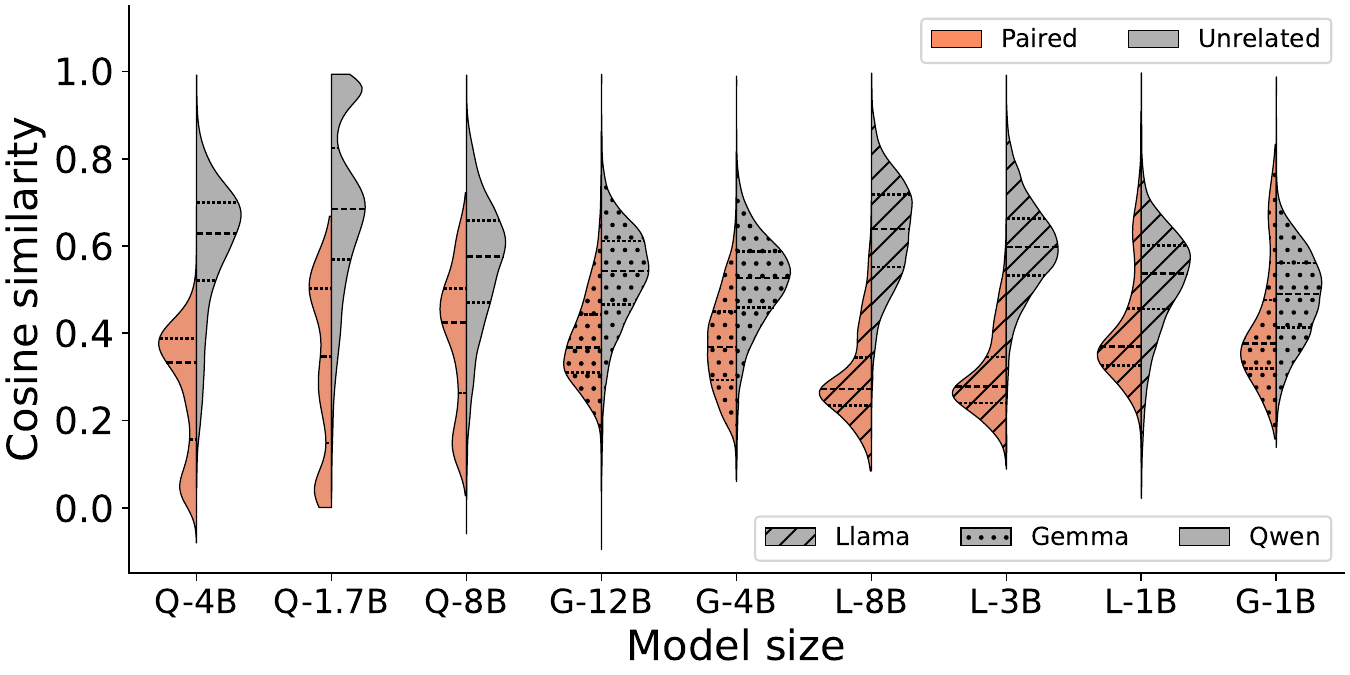}
\caption{\textbf{Dialectal form outweighs semantic mismatch in gradient geometry.} Violin plots of the gradient cosine similarity between paired and unrelated documents on \textsc{ReDial} reveal that matched SAE-AAVE pairs produce lower similarity than unrelated SAE-SAE pairs, implying that a meaning-preserving dialect shift can perturb the training signal more than changing the document content within SAE. In the plot, an LM name is indicated by its first-letter abbreviation and size.}
\label{figure:gradient_analysis-paired_vs_unrelated}
\end{figure}

\noindent\textbf{Matched texts yield mismatched gradients.} We compare two distributions: (i) the paired similarity between SAE and AAVE versions of the same text and (ii) the unpaired similarity between unrelated SAE texts. Under the null hypothesis that the model is dialect-invariant, dialect pairs that share semantic meaning should induce gradient updates more similar than those of unrelated documents (i.e., $\mu^+ \gg \mu^-$). Figure~\ref{figure:gradient_analysis-paired_vs_unrelated} rejects this hypothesis. Across all models and tasks, $\mu^+ < \mu^-$ with a mean $z = -2.64$, indicating that the dialect transformation disrupts gradient geometry more than substituting an entirely different document. Concretely, the gradient induced by the AAVE encoding of problem $i$ has lower cosine similarity to that of its SAE counterpart than to that of an unrelated SAE sample $j$. A pre-trained base model produces more divergent parameter updates on semantically identical dialectal content than on semantically unrelated SAE content, pointing to disparities in learning costs between different dialectal texts.

\vspace{.5\baselineskip}

\begin{figure}[b]
\centering
\includegraphics[width=\linewidth]{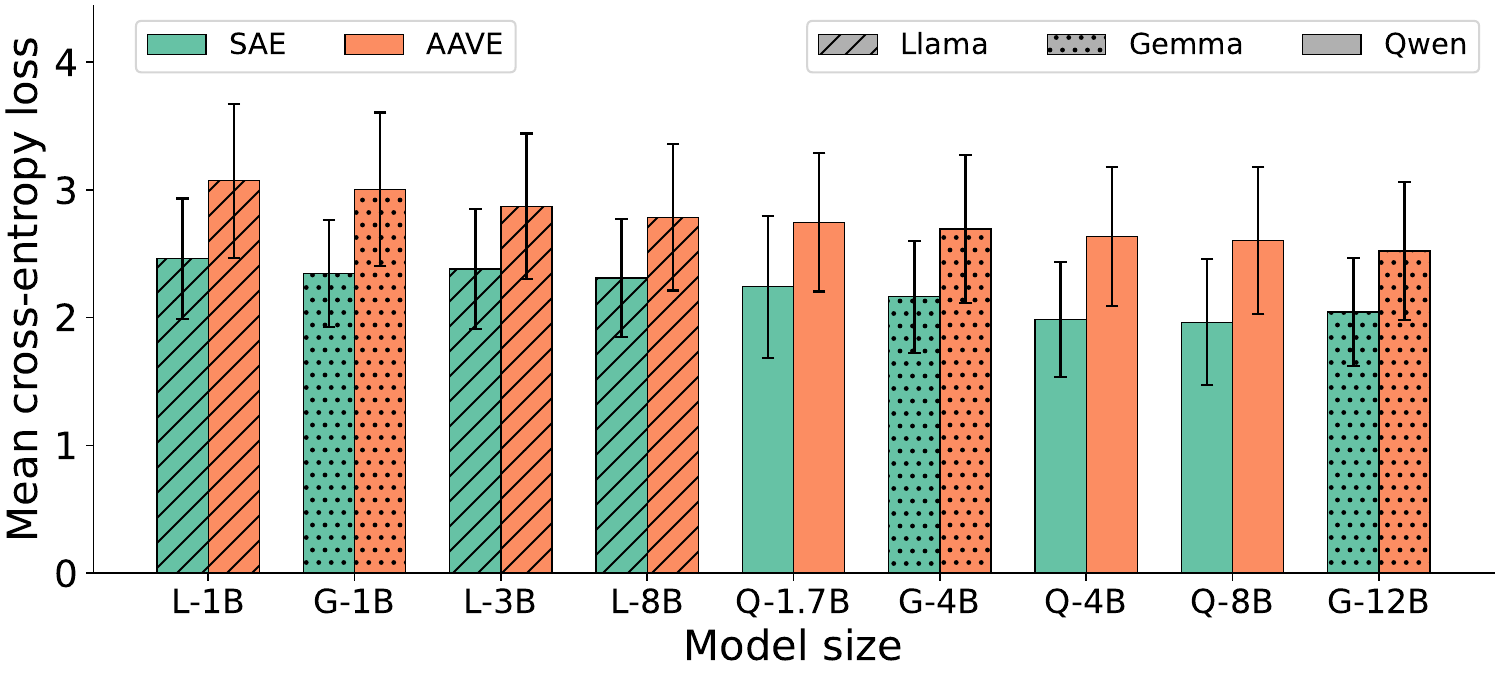}
\caption{\textbf{Dialectal text incurs higher prediction loss.} We plot the mean cross-entropy loss by dialect. All nine models assign significantly higher loss to AAVE inputs than to their SAE counterparts (one-sided paired Wilcoxon test, $n=1{,}200$ per model, Bonferroni-corrected $p<0.001$), with per-model mean gaps of $0.47$ to $0.66$ nats (Cohen's $d_z \in [2.12, 2.55]$). An LM name is indicated by its first-letter abbreviation and size.}
\label{figure:gradient_analysis-cross_entropy_loss}
\end{figure}

\noindent\textbf{Loss gaps persist regardless of correctness.} The dialect loss gap is invariant to model family and scale, resulting in inflated per-token prediction cost on dialectal text. In Figure~\ref{figure:gradient_analysis-cross_entropy_loss}, all nine models assign consistently higher cross-entropy loss to AAVE inputs than to their SAE counterparts, with a mean gap of approximately $0.56$ nats. Models find AAVE text systematically harder to predict than SAE text.

To rule out problem difficulty as a confound, we calculate the point-biserial correlation between $s_i^+$ and a binary indicator for correct answers on both dialects. Indeed, the correlation pooled across all models is negligible ($r = -0.013$, $p = 0.19$, $n = 10{,}800$), and the correlation per-model is weak ($|r| \le 0.19$), as listed in Table~\ref{table:gradient_similarity_correctness_correlation}. The conditional expectations $\mathbb{E}[s_i^+ | \text{both correct}] = 0.348$ and $\mathbb{E}[s_i^+ | \text{any incorrect}] = 0.354$ differ by $0.006$ nats, which is much smaller than the dialect gap itself. This result falsifies the hypothesis that gradient divergence concentrates on problems the model finds difficult, meaning parameter updates are distorted uniformly regardless of correctness. Thus, improvements in model accuracy alone fail to mitigate the resulting dialect bias from pre-training.

\vspace{.5\baselineskip}

\noindent\textbf{Dialectal texts rival character noise.} Naturally, one might object that any transformed text yields noisier gradients than its matched SAE text. We calibrate the AAVE $z$-score against five character-level perturbations\footnote{We use swap, drop, insert, random capitalization, and alternating capitalization.} of SAE from Section~\ref{same_meaning_different_taxes}, recomputing each $z = (\mu^+ - \mu^-)/\sigma^-$ where $\mu^+$ is now the paired similarity between SAE and its perturbed version, against the shared unrelated-SAE baseline ($\mu^- = 0.57$, $\sigma^- = 0.10$). Table~\ref{table:gradient_perturbation_calibration} compares the AAVE-SAE divergence to each perturbation, showing AAVE is more disruptive than character swaps or insertions and sits on the order of dropping $15\%$ of characters. Given that AAVE preserves meaning whereas character perturbations can corrupt it, our results suggest that dialectal surface variation elicits gradient divergence comparable to that induced by substantial character-level corruption.

\subsection{Post-Training Rewards}
\label{dialect_taxes_posttraining}

A critical component of LM alignment is the reward model (RM), which provides the optimization signal for post-training. Biases encoded by the RM propagate into the policy of the deployed LM. Here, we measure rewards on parallel SAE-AAVE content to identify systematic dialect preferences.

\vspace{.5\baselineskip}

\noindent\textbf{Setup.} We evaluate ten RMs from three providers (\textsc{Ai2}, \textsc{QRM}, \textsc{Skywork}) spanning 3B to 70B parameters across \textsc{Gemma-2}, \textsc{Llama-3}, and \textsc{Qwen-3} LM families (\S\ref{appendix:reward_models}). To obtain reward scores $r(x, y)$ for an input $x$ and response $y$, we calculate (1) sample-level scores and (2) token-level scores.

For sample-level scoring, we use the \textsc{ReDial} dataset to construct a prompt that concatenates a reasoning problem $x_i$ with its gold-standard answer $y_i$, \texttt{``\{problem\}\{answer\}''}, such that each chosen-rejected preference pair presents the same question with the same answer as the assistant response in both dialects. To measure an RM preference for SAE text over AAVE text, we define $\Delta r_i = r(x_i^{\text{SAE}}, y_i) - r(x_i^{\text{AAVE}}, y_i)$.

For token-level scoring, we use \textsc{ReDial}, \textsc{ParallelAAVE}, and \textsc{MultiVALUE} to identify subword tokens that appear exclusively in tokenized dialect texts (e.g., ``wanna'', ``lil'') and SAE texts (e.g., ``Calculate'', ``regardless'') across the seven tokenizers from Section~\ref{dialectal_biases}. We use the method by \citet{christian2026rewardbiases} to isolate the rewards of each token from any prior context. Given a fixed prompt $x$, \enquote{What, in one word or subword, is the greatest thing ever?}, we present each token individually as the response $y$ to receive score $r(x, y)$.

\begin{table}[t]
\centering
\small
\begin{tabular}{lcc}
\toprule
\textbf{Paired condition with SAE} & $\mu^+ (\uparrow)$ & $z$ \\
\midrule
Baseline (unrelated SAE)   & $0.57$ & $0.00$ \\
\noalign{\vskip\aboverulesep}\cdashline{1-3}\noalign{\vskip\belowrulesep}
Capitalize (alternating)     & $0.29$ & $-3.36$ \\
Capitalize (random)          & $0.34$ & $-2.88$ \\
Drop ($\mathbb{P} = 0.15$)   & $0.36$ & $-2.58$ \\
Insert ($\mathbb{P} = 0.05$)  & $0.47$ & $-1.40$ \\
Swap ($\mathbb{P} = 0.05$)    & $0.50$ & $-1.06$ \\
\noalign{\vskip\aboverulesep}\cdashline{1-3}\noalign{\vskip\belowrulesep}
Dialect (AAVE)                & $0.35$ & $-2.64$ \\
\bottomrule
\end{tabular}
\caption{\textbf{Dialect gradient divergence sits within the range of character-level perturbations.} Across nine base models and four \textsc{ReDial} tasks, we report the grand-mean paired cosine similarity ($\mu^+$) and $z$-score against the unrelated-SAE baseline. Higher $\mu^+$ indicates greater similarity to SAE gradients; lower values indicate stronger gradient divergence.}
\label{table:gradient_perturbation_calibration}
\end{table}

\vspace{.5\baselineskip}

\noindent\textbf{Reward models have incoherent dialect bias.} If RMs scored only on semantic content, each paired SAE-AAVE reward gap should satisfy $\Delta r_i = 0$, and therefore $\mathbb{E}[\Delta r] = 0$. We test this null hypothesis with a one-sample $t$-test over paired reward gaps. Eight RMs reject this null at $p < 0.05$, except \textsc{Skywork}-\textsc{Llama}~3B ($p = 0.50$) and \textsc{Skywork}-\textsc{Gemma}~27B ($p = 0.76$). Thus, most RMs carry some directional dialect signal, but two patterns preclude the simplest explanations of this gap. First, the bias is not a uniform AAVE penalty, as the $\Delta r$ sign flips across \textsc{ReDial} tasks. Pooled across RMs, Algorithm ($\overline{\Delta r} = +0.43$) and Math ($+0.04$) lean SAE while Logic ($-0.08$) and Planning ($-0.14$) lean AAVE. While all tasks are significant ($p < 0.01$) but Math ($p = 0.239$), effect sizes are uniformly small (Cohen's $|d| \leq 0.25$). That is, every RM flips dialect preference by topic. Second, the bias is not a fixed artifact of the pre-trained backbone. Within \textsc{Ai2}, the mean gap flips from $+0.41$ for the base \textsc{Llama}~8B to $-0.19$ after instruction tuning, reversing rather than removing the dialect tax. Taken together, these dialectal differences are neither uniform across content nor stable across training stages, rendering RM preferences too unstable to correct pre-training dialectal gaps.

\vspace{.5\baselineskip}

\noindent\textbf{Rewards are contextual, not lexical.} A natural explanation for the sample-level dialect gap is that the bias lives in the surface tokens, that RMs penalize dialect-exclusive subwords. Token-level scoring reveals the opposite: dialect-exclusive tokens receive higher scores than SAE-exclusive tokens across all corpora, tokenizers, and RMs (pooled mean score gap $\bar{r}_{\text{SAE}} - \bar{r}_{\text{dialect}} = -0.55$, $p < 0.001$ with an independent $t$-test)\footnote{Reward scales differ substantially across RMs, so pooled gaps are not model-balanced. We normalize this in Table~\ref{table:rewards_token_per_rm_dialect}.}. The direction is consistent across each corpora (Table~\ref{table:rewards_token_by_corpus}) and statistically significant with $p < 0.001$ under a pooled two-sample $t$-test across all five dialects after RM scaling (Table~\ref{table:rewards_token_by_dialect_scaling}). Our results highlight that isolated-token and full-context preferences can be decoupled in RMs, which reinforce findings that preference-tuned LMs can sometimes exhibit covert, without overt, prejudice against AAVE \citep{hofmann2024aicovertracism}. Although alignment ideally corrects undesirable pre-training biases, these RMs fail to provide consistent corrective signals for preference learning to close pre-trained dialect gaps.

\subsection{Inference-Time Representations}
\label{dialect_taxes_inference}

We ask whether the dialectal biases traced above are visible during inference along two complementary axes: hidden states and output distributions.

\vspace{.5\baselineskip}

\noindent\textbf{Setup.} We probe nine models across three LM families (\textsc{Llama-3}: 1B, 3B, 8B; \textsc{Gemma-3}: 1B, 4B, 12B; \textsc{Qwen-3}: 1.7B, 4B, 8B). (1) \emph{Hidden states.} Using both the base and instruct model variants, we extract the mean-pooled hidden state $h_\ell \in \mathbb{R}^{d_\text{model}}$ at every layer $\ell \in \{0,\ldots,L\}$ for each parallel pair in \textsc{MultiVALUE} and \textsc{ParallelAAVE}. We then compute $s_\ell = \text{sim}(h_\ell^{\text{SAE}}, h_\ell^{\text{dialect}})$ to measure the similarity of paired representations at each layer. (2) \emph{Output distributions.} Using the instruct models, we compare AAVE inputs on \textsc{ReDial} against the five SAE character perturbations from Section~\ref{same_meaning_different_taxes} by measuring their input cross-entropy, generation entropy, and answer accuracy. These perturbations calibrate AAVE against surface-form noise.

\vspace{.5\baselineskip}

\begin{figure}[b]
\centering
\includegraphics[width=\linewidth]{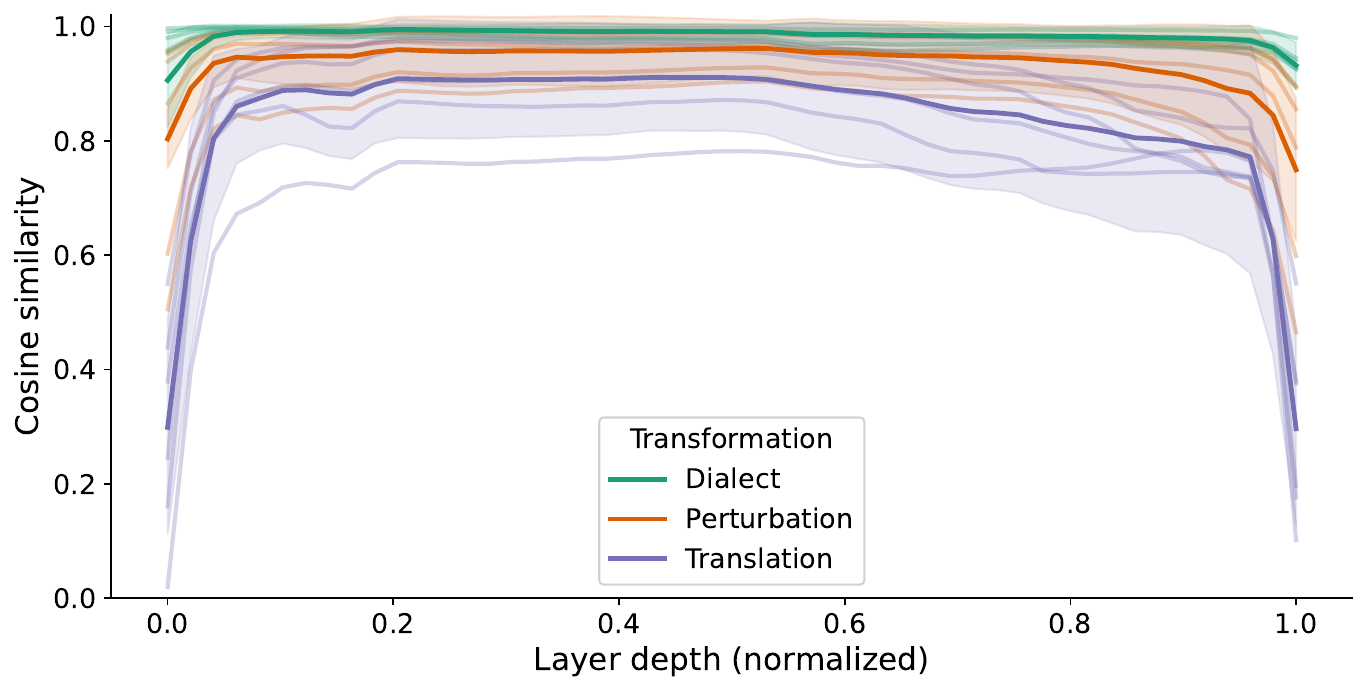}
\caption{\textbf{Transformations result in similar hidden-state similarity curves.} Layer-wise cosine similarities between SAE hidden states and those of transformed texts are pooled across \textsc{MultiVALUE} and \textsc{ParallelAAVE}. All text transformations follow similar trajectories across normalized LM layers, with dialectal transformations most similar to SAE.}
\label{figure:hidden_state_group_comparison}
\end{figure}

\noindent\textbf{Dialects look similar to noise.} All transformations produce similar hidden-layer profiles, as shown in Figure~\ref{figure:hidden_state_group_comparison}, that exhibit low similarity at the embedding layer due to differences in surface-form tokenizations, experience rapid convergence in similarity in intermediate layers, then diverge near the output as representations differentiate in next-token prediction. We verify that across all $504$ model $\times$ dataset $\times$ transformation cells, Friedman $\chi^2$ tests reject flat layer profiles with $p < 0.001$, and for almost every ($502$ of $504$) condition, final-layer similarity drops significantly below its peak (one-sided Wilcoxon signed-rank test, $p < 0.001$). The magnitude of the final-layer similarities differs between transformation types. In paired comparisons across $36$ model $\times$ dataset combinations, final-layer similarity for dialectal transformations is significantly higher than that for $10$ of $11$ comparison transformations (one-sided Wilcoxon signed-rank, Bonferroni-corrected $p < 0.01$).

In terms of entropy, dialectal transformations resemble character noise. Compared to SAE, AAVE inputs increase next-token entropy by $+0.29$ nats on average, within the range for noise-perturbed text (mean $+0.22$ nats), whereas translations produce little change (mean $-0.03$ nats). Generation cross-entropy remains largely unchanged under every condition ($|\Delta| < 0.03$ nats), precluding degenerate outputs for all tested text transformations.

\vspace{.5\baselineskip}

\begin{figure}[t]
\centering
\includegraphics[width=\linewidth]{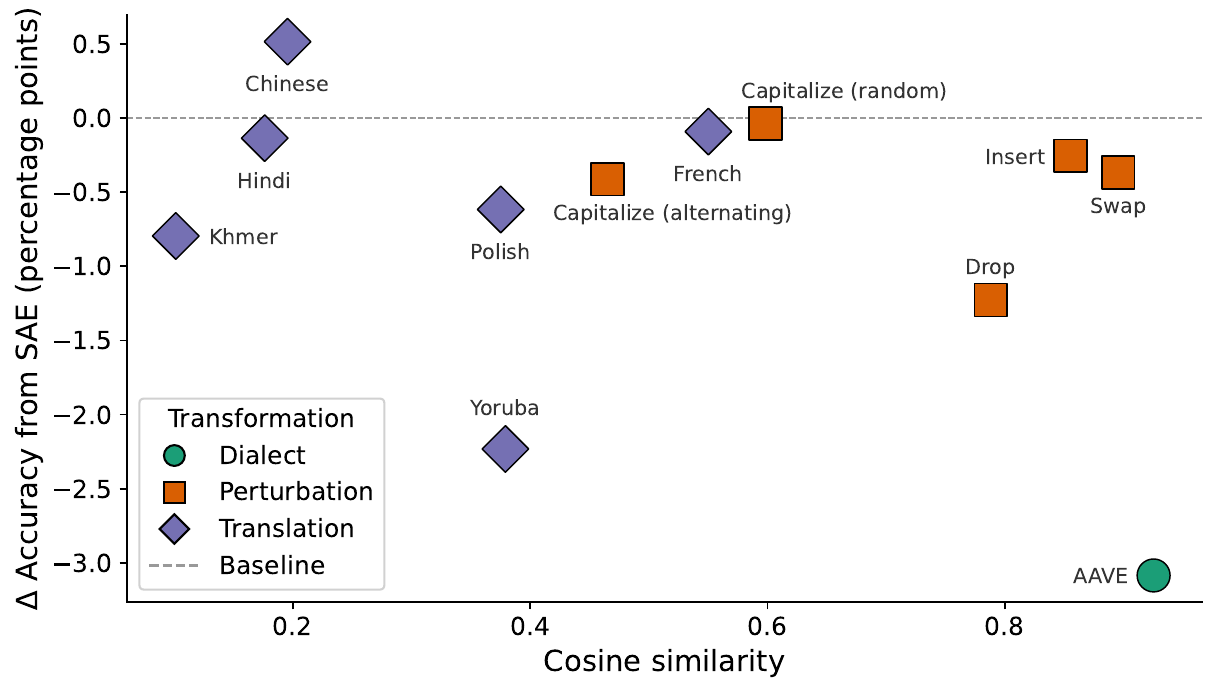}
\caption{\textbf{Hidden-state similarity $\centernot\Rightarrow$ downstream accuracy.} On the \textsc{ReDial} dataset, we plot each transformation's final-layer hidden-state cosine similarity to SAE against its change in answer accuracy. Among semantic-meaning-preserving transformations, SAE-to-AAVE shows the highest similarity to the original SAE representation but the lowest downstream accuracy.}
\label{figure:hidden_state_similarity_vs_accuracy}
\end{figure}

\noindent\textbf{Models perform worse on dialect than noise.} Despite comparable internal inference-time representations, AAVE results in greater accuracy loss ($3.09$ pp) than noise and translation transformations ($0.04$ to $1.23$ pp) on \textsc{ReDial} (Figure~\ref{figure:hidden_state_similarity_vs_accuracy}). Ultimately, dialectal text may yield representational profiles similar to those of generic noise yet produce worse downstream outcomes, levying ``the dialect tax''.

\section{Discussion}
\label{discussion}

The key finding of this work is that the dialect tax is introduced at every stage of the language modeling pipeline. To our knowledge, this paper provides the first comprehensive study of dialect discrepancies across each component of an LM. Prior research on dialectal biases in models have examined individual language modeling components in isolation, often either surfacing the existence of a gap without identifying its source or proposing post-hoc interventions for biases already encoded in an LM. We argue that the dialect tax manifests in multiple locations and that pinpointing a singular source of dialect prejudice obscures the need to address the issue holistically. Notably, we present evidence that individual interventions mitigate specific measured disparities without resolving the underlying gap. To avoid the problem illustrated by the parable of the blind men and the elephant, proposed interventions should be evaluated for their validity across the entire model pipeline. Rather than implying that the language modeling process is prejudiced, our findings highlight the necessity of techniques that account for the systemic influence of linguistic variation throughout the model pipeline.

%%%%%%%%%%%%%%%%%%%%%%%%%%%%%%

\section*{Limitations}

Our work aims to study the sources of dialectal bias within each stage of the language modeling pipeline, but our experiments are necessarily limited by our computational resources and techniques. Below, we cover broad limitations of this study.

\vspace{.5\baselineskip}

\noindent\textbf{Datasets.} Our study is empirical in nature, and as such, we needed to narrow our experimental setting to prioritize controlled parallel comparisons of standard versus non-standard English dialects. The best publicly-available resources we found differed in their construction processes. \textsc{ParallelAAVE} contains human rewrites of tweets and therefore emphasizes conversational online language found on social media, \textsc{ReDial} focuses on popular reasoning benchmarks written in more formal academic language, and \textsc{MultiVALUE} uses a rule-based system to generate dialectal text instead of eliciting it directly from speakers. These datasets allow us to fix semantic meaning while varying dialectal surface form, which, while central to our claims, do not capture the situational range in which the dialect tax may appear. A next step is to build and evaluate community-validated parallel corpora across more dialects and settings, making sure to collect data from native speakers rather than generating text using translation rules.

\vspace{.5\baselineskip}

\noindent\textbf{Models.} We host smaller open-source LMs (\textsc{Gemma-3}, \textsc{Llama-3}, \textsc{Qwen-3}) and RMs (\textsc{Ai2}, \textsc{QRM}, and \textsc{Skywork}) because their model internals (e.g., tokenizers, losses, gradients, hidden states, and reward scores) can be inspected under our computational budget. However, closed commercial systems may exhibit different behaviors, as their training pipelines may amplify or mitigate dialectal disparities in ways our methods cannot verify without internal access. We are further limited by our compute and budget constraints, which hinder our ability to repeat every experiment across multiple random seeds. Instead of measuring statistical significance across different runs, our statistical evidence relies on paired-sample variation across different inputs and models. We leave the study of dialect tax in larger models to future work.

\vspace{.5\baselineskip}

\noindent\textbf{Interventions.} Our interventions diagnose mechanisms more than they prescribe a remedy. Experiments in this work are primarily observational, exposing behavioral differences to dialects at inference time but lacking controlled ablations for model training. We conduct no model training, meaning our experiments cannot identify the point at which dialectal disparities become encoded during pre-training or post-training. The intended continuation of our work aims to identify the causal mechanisms of the dialect tax through controlled experiments that explicitly test whether individual components of the language modeling pipeline preserve dialectal equivalence.

%%%%%%%%%%%%%%%%%%%%%%%%%%%%%%

\section*{Ethics Statement}

We abide by the general principles of research in the NLP community. To ensure safety and privacy, our study used open-source datasets whose data was collected with informed consent and pseudonymized participant identities.

%%%%%%%%%%%%%%%%%%%%%%%%%%%%%%

\section*{Acknowledgments}

We thank the anonymous reviewers who provided feedback for this paper. We are grateful to Valentin Hofmann and Martijn Bartelds for guidance, particularly on dataset selection, linguistic variation, and prior literature. Our gratitude extends to Joshua Kazdan and Harsha Nori for insightful discussions.

%%%%%%%%%%%%%%%%%%%%%%%%%%%%%%

\bibliography{custom}

%%%%%%%%%%%%%%%%%%%%%%%%%%%%%%

%%%%%%%%%%%%
% APPENDIX %
%%%%%%%%%%%%

\appendix
\tableofcontents

\section{Data}
%%%%%%%%
% DATA %
%%%%%%%%

We open source our code on GitHub at the following link:
\begin{center}
\url{https://github.com/socialnlp/dialecttax}
\end{center}

\subsection{\textsc{ReDial}}
\label{appendix:redial}

Our experiments rely on the \textsc{ReDial} \citep{redial} dataset, which is a reasoning benchmark containing parallel query pairs in SAE and AAVE. We reconstructed \textsc{ReDial} from their original seven source datasets, which are summarized in Table~\ref{table:data-redial} and detailed in the sections below. This allowed us to have greater control in modifying the natural language prompts within our experiments.

In general, we attempted to unify the format of the dataset as much as possible. For instances that conflict between \textsc{ReDial} and the source datasets, we defer to the original source.

We note that there are many small inconsistencies, including punctuation, capitalization, or\footnote{Inclusive \enquote{or}, i.e. $\lor$.} grammar mistakes that originated from the dataset sources or were introduced by \textsc{ReDial}. We corrected, to our discretion, many of these inconsistencies (described within individual source sections) and provided metadata labels to the original sources. We also remove duplicated samples.

For the experiments presented in the main section, we use our modified re-constructed \textsc{ReDial} dataset formatted a multiple-choice question and answer dataset.

\begin{table*}[ht]
\centering
\begin{tabular}{llr}
\toprule
\textbf{Category} & \textbf{Source} & \textbf{Items} \\
\midrule
\multirow{2}{*}{\textbf{Algorithm} (26\%)} & \href{https://github.com/openai/human-eval}{\textsc{HumanEval}} \citep{humaneval} & 164 \\
                           & \href{https://github.com/google-research/google-research/blob/master/mbpp/sanitized-mbpp.json}{\textsc{MBPP} (Sanitized)} \citep{mbpp} & 149 \\
\midrule
\multirow{2}{*}{\textbf{Logic} (30\%)} & \href{https://huggingface.co/datasets/yale-nlp/FOLIO}{\textsc{FOLIO}} \citep{folio,folioperturbed} & 162 \\
                       & \href{https://github.com/Mihir3009/LogicBench}{\textsc{LogicBench}} \citep{logicbench} & 200 \\
\midrule
\multirow{2}{*}{\textbf{Math} (25\%)} & \href{https://huggingface.co/datasets/openai/gsm8k}{\textsc{GSM8K}} \citep{gsm8k} & 150 \\
                      & \href{https://huggingface.co/datasets/ChilleD/SVAMP}{\textsc{SVAMP}} \citep{svamp} & 150 \\
\midrule
\textbf{Planning} (19\%) & \href{https://huggingface.co/datasets/fangrulin/asynchow}{\textsc{AsyncHow}} \citep{asynchow} & 225 \\
\midrule
\textbf{Total} & - & 1,200 \\
\bottomrule
\end{tabular}
\caption{\textbf{Source datasets used to reconstruct \textsc{ReDial}.}}
\label{table:data-redial}
\end{table*}

\subsubsection{Algorithm}

We remove \textsc{ReDial} prompt instructions to be added programmatically during inference.

In accordance with \textsc{ReDial}, we substitute all function names as \texttt{python\_function} to avoid as much memorization as possible. We corrected several corrupted questions where \textsc{ReDial} incorrectly replaced the original function keyword with \enquote{python\_function} (e.g. we corrected \enquote{Write a function to find the nth number in the newman conway python\_function.} in \textsc{ReDial} to \enquote{Write a function to find the nth number in the newman conway sequence.}).

\vspace{.5\baselineskip}

\noindent\textbf{\textsc{HumanEval}.} The dataset includes $164$ programming problems released by OpenAI. We reference the following hyperlinked \href{https://github.com/openai/human-eval}{code} and \href{https://huggingface.co/datasets/openai/openai_humaneval}{dataset}.\footnote{The \textsc{ReDial} dataset used the instruction-following queries available within \textsc{InstructHumanEval} at \url{https://huggingface.co/datasets/codeparrot/instructhumaneval}.} Although \textsc{ReDial} modified the instructions of some samples within their data, we use the original dataset released by OpenAI.

\vspace{.5\baselineskip}

\noindent\textbf{\textsc{MBPP}.} The \textsc{MBPP} dataset, a.k.a. the \enquote{Mostly Basic Python Problems} dataset, is a benchmark of crowd-sourced Python programming problems designed to be solvable by entry-level programmers. We reference the following hyperlinked \href{https://github.com/google-research/google-research/tree/master/mbpp}{dataset}, specifically the hand-verified \enquote{sanitized} verison. Because \textsc{MBPP} did not include the context included in \textsc{HumanEval}, we generated the context using \textsc{GPT-5 Mini} through OpenRouter\footnote{\url{https://openrouter.ai/}} and verify the results manually. We also removed one duplicated sample within \textsc{ReDial}.

\subsubsection{Logic}

\noindent\textbf{\textsc{FOLIO}.} The original \textsc{FOLIO} dataset is a manually curated logic benchmark. The counterfactual dataset was created to explore the capabilities of language models through counterfactual evaluations on various domains, such as logic. Since the perturbed counterfactual dataset contains both the original and perturbed versions, we reference the following hyperlinked \href{https://github.com/ZhaofengWu/counterfactual-evaluation/tree/master/logic}{code} and \href{https://huggingface.co/datasets/ZhaofengWu/FOLIO-counterfactual/blob/main/folio_v2_perturbed.jsonl}{dataset}.

\vspace{.5\baselineskip}

\noindent\textbf{\textsc{LogicBench}.} The dataset offers a natural language question-answering evaluation for the logical reasoning abilities of large language models. We reference the following hyperlinked \href{https://github.com/Mihir3009/LogicBench}{dataset}. To the best of our ability, we capitalized proper nouns and the first letter of sentences to prevent confounders from non-standard grammar. The \textsc{ReDial} dataset seems to have permuted the multiple choice answers, so we maintain this new ordering. We also kept the changes from \textsc{ReDial} that corrected incorrect names from the original data source. See our code for more details.

\subsubsection{Math}

\noindent\textbf{\textsc{GSM8K}.} The dataset consists of grade-school math word problems with $8,790$ instances. We reference the following hyperlinked \href{https://github.com/openai/grade-school-math}{code} and \href{https://huggingface.co/datasets/openai/gsm8k}{dataset}. We matched each \textsc{ReDial} instance to a \textsc{GSM8K} instance and used the original data.

\vspace{.5\baselineskip}

\noindent\textbf{\textsc{SVAMP}.} The dataset contains $1,000$ elementary-school math word problems. We reference the following hyperlinked \href{https://github.com/arkilpatel/SVAMP}{code} and \href{https://huggingface.co/datasets/ChilleD/SVAMP}{dataset}. In keeping with \textsc{ReDial}, we use only the \textsc{SVAMP} test set. To avoid unnecessary confounders in our dataset, we used \textsc{GPT-5 Mini} through OpenRouter to fix the grammar of the questions. We manually verify the generated results.

\subsubsection{Planning}

\noindent\textbf{\textsc{AsyncHow}.} The dataset contains benchmarks for planning reasoning tasks. We reference the following hyperlinked \href{https://github.com/fangru-lin/graph-llm-asynchow-plan}{code} and \href{https://huggingface.co/datasets/fangrulin/asynchow}{dataset}. We removed 15 duplicates from the original \textsc{ReDial} dataset for a total of 225 instances. Additionally, there were several inconsistencies between the original \textsc{AsyncHow} and the \textsc{ReDial} datasets, e.g. missing spaces and changed characters. For any discrepancies, we defer to the original \textsc{AsyncHow} dataset. We also change the following:

\begin{itemize}[nolistsep]
    \item Change instructions from \enquote{These ordering constraints need to be obeyed when executing above steps} to \enquote{These ordering constraints must be followed when executing the above steps}
    \item Correct the capitalization of the first line description
    \item Correct the capitalization of label \enquote{step <\#>} to \enquote{Step <\#>}
    \item Change \enquote{Step <\#>. <description>} to the more natural \enquote{Step <\#>: <description>}
    \item Remove extraneous \enquote{\textbackslash n\textbackslash n} from \textsc{ReDial} not present in \textsc{AsyncHow}
\end{itemize}

It is important to note that we've made the task substantially more difficult by enforcing the models to answer in seconds. This makes the task much more readily verifiable, although it is still possible to analyze the results for non-compliant completions (e.g. \enquote{3 days 7 hours}) using fuzzy matching.

\subsection{\textsc{ParallelAAVE} and \textsc{MultiVALUE}}

Table~\ref{table:data-parallelaave-multivalue} summarizes the dialects included in our parallel text datasets.

\begin{table*}[ht]
\centering
\resizebox{\textwidth}{!}{
\begin{tabular}{llcl}
    \toprule
    \textbf{Dataset} & \textbf{Source} & \textbf{Pairs} & \textbf{Dialects}\\
    \midrule
    \textsc{ParallelAAVE} & \citet{parallelaave} & $2019$ & SAE, AAVE \\
    \textsc{MultiVALUE} & \citet{multivalue} & $429$ & SAE, AAVE, Appalachian, Chicano, Indian, Singapore\\
    \bottomrule
\end{tabular}}
\caption{\textbf{Datasets for parallel SAE and AAVE texts.}}
\label{table:data-parallelaave-multivalue}
\end{table*}

\subsection{Licenses}

\textsc{ReDial} is held under the MIT License. \textsc{MultiVALUE} is held under the Apache License, Version 2.0. \textsc{ParallelAAVE} did not specify the license for their data, but it was published by the Association for Computational Linguistics in 2020, which defaults to the Creative Commons Attribution 4.0 International (CC BY 4.0) license.

\section{Models}
%%%%%%%%%%
% MODELS %
%%%%%%%%%%

\subsection{Tokenizers}

\subsubsection{Tokenization Algorithms}

\noindent\textbf{\textsc{BPE}.} Based on the byte-pair encoding (\textsc{BPE}) compression algorithm, \citet{sennrich2015bpe} introduced this tokenization approach to aid in open-vocabulary machine translation tasks by encoding rare or unknown words into sequences of subword units. It is a simple technique that iteratively merges frequent pairs of bytes or byte sequences into one sequence. It is the common approach used by LM families, such as \textsc{GPT} \citep{radford2018gpt}, \textsc{Gemma} \citep{google2025gemma3}, \textsc{Llama} \citep{grattafiori2024llama3herdmodels}, and \textsc{Qwen} \citep{yang2025qwen3technicalreport}.

\vspace{.5\baselineskip}

\noindent\textbf{\textsc{Unigram}.} The algorithm is initialized with a large base vocabulary, which is iteratively trimmed to minimize the overall loss over the training data \citep{kudo2018unigram}. It is used in combination with the \textsc{SentencePiece} tokenizer  \citep{kudo2018sentencepiece} used in LMs, like \textsc{T5} \citep{raffel2023t5}.

\vspace{.5\baselineskip}

\noindent\textbf{\textsc{WordPiece}.} Similar to \textsc{BPE}, \textsc{WordPiece} iteratively merges pairs of bytes or byte sequences into one sequence that maximizes the likelihood of the training data \citep{schuster2012wordpiece}. It is the tokenization algorithm used in \textsc{BERT} \citep{devlin2018bert}.

\subsubsection{Tokenizer Models}

Table~\ref{table:tokenizers} lists the tokenizer names used in our analysis. We use the tokenizer corresponding to the largest size of a particular LM family, although this distinction does not matter.

\begin{table}[H]
\centering
\resizebox{\linewidth}{!}{
\begin{tabular}{lll}
    \toprule
    \textbf{Tokenization} & \textbf{Name} & \textbf{Model / Encoding}\\
    \midrule
    \multirow{5}{*}{\textsc{BPE}} & \textsc{GPT-5} & \texttt{o200k\_base} (tiktoken)\\
    & \textsc{GPT-2} & \texttt{openai-community/gpt2}\\
    & \textsc{Gemma} & \texttt{google/gemma-3-27b-it}\\
    & \textsc{Llama} & \texttt{meta-llama/Llama-3.3-70B-Instruct}\\
    & \textsc{Qwen} & \texttt{Qwen/Qwen3-32B}\\
    \midrule
    \textsc{Unigram} & T5 & \texttt{t5-small}\\
    \midrule
    \textsc{WordPiece} & BERT & \texttt{bert-base-uncased}\\
    \bottomrule
\end{tabular}
}
\caption{\textbf{Tokenizer names used in our analysis.}}
\label{table:tokenizers}
\end{table}

\subsection{Language Models}
\label{appendix:language_models}

Table~\ref{table:language-models} lists all LMs used in our analysis, including their Hugging Face model IDs.\footnote{Note that there is no base model version of \textsc{Qwen-3 32B}.}

\begin{table*}[b]
\centering
\resizebox{.75\linewidth}{!}{
\begin{tabular}{lllll}
    \toprule
    \textbf{Family} & \textbf{Name} & \textbf{Size} & \textbf{Type} & \textbf{Model ID}\\
    \midrule
    \multirow{6}{*}{\textsc{Llama}} & \textsc{Llama 3.2} & 1B & Base & \texttt{meta-llama/Llama-3.2-1B}\\
    & \textsc{Llama-3.2} & 1B & Instruct & \texttt{meta-llama/Llama-3.2-1B-Instruct}\\
    & \textsc{Llama-3.2} & 3B & Base & \texttt{meta-llama/Llama-3.2-3B}\\
    & \textsc{Llama-3.2} & 3B & Instruct & \texttt{meta-llama/Llama-3.2-3B-Instruct}\\
    & \textsc{Llama-3.1} & 8B & Base & \texttt{meta-llama/Llama-3.1-8B}\\
    & \textsc{Llama-3.1} & 8B & Instruct & \texttt{meta-llama/Llama-3.1-8B-Instruct}\\
    & \textsc{Llama-3.1} & 70B & Base & \texttt{meta-llama/Llama-3.1-70B}\\
    & \textsc{Llama-3.1} & 70B & Instruct & \texttt{meta-llama/Llama-3.1-70B-Instruct}\\
    \midrule
    \multirow{6}{*}{\textsc{Gemma}} & \textsc{Gemma 3} & 1B & Base & \texttt{google/gemma-3-1b-pt}\\
    & \textsc{Gemma-3} & 1B & Instruct & \texttt{google/gemma-3-1b-it}\\
    & \textsc{Gemma-3} & 4B & Base & \texttt{google/gemma-3-4b-pt}\\
    & \textsc{Gemma-3} & 4B & Instruct & \texttt{google/gemma-3-4b-it}\\
    & \textsc{Gemma-3} & 12B & Base & \texttt{google/gemma-3-12b-pt}\\
    & \textsc{Gemma-3} & 12B & Instruct & \texttt{google/gemma-3-12b-it}\\
    & \textsc{Gemma-3} & 27B & Base & \texttt{google/gemma-3-27b-pt}\\
    & \textsc{Gemma-3} & 27B & Instruct & \texttt{google/gemma-3-27b-it}\\
    \midrule
    \multirow{6}{*}{\textsc{Qwen}} & \textsc{Qwen 3} & 1.7B & Base & \texttt{Qwen/Qwen3-1.7B-Base}\\
    & \textsc{Qwen-3} & 1.7B & Instruct & \texttt{Qwen/Qwen3-1.7B}\\
    & \textsc{Qwen-3} & 4B & Base & \texttt{Qwen/Qwen3-4B-Base}\\
    & \textsc{Qwen-3} & 4B & Instruct & \texttt{Qwen/Qwen3-4B}\\
    & \textsc{Qwen-3} & 8B & Base & \texttt{Qwen/Qwen3-8B-Base}\\
    & \textsc{Qwen-3} & 8B & Instruct & \texttt{Qwen/Qwen3-8B}\\
    & \textsc{Qwen-3} & 32B & Instruct & \texttt{Qwen/Qwen3-32B}\\
    \bottomrule
\end{tabular}
}
\caption{\textbf{Language models used in our analysis.}}
\label{table:language-models}
\end{table*}

\subsection{Embedding Models}

We use Google's open embedding model \textsc{EmbeddingGemma-300M} (Hugging Face Model ID: \texttt{google/embeddinggemma-300m}) built from \textsc{Gemma-3} to produce vector representations of text.

As suggested on the \textsc{EmbeddingGemma} model page, we tailor our similarity analysis query to generate embeddings that are optimized to assess text similarity by prefixing the content with the following prompt: \enquote{\texttt{task: sentence similarity | query: \{content\}}}.

\subsection{Reward Models}
\label{appendix:reward_models}

Table~\ref{table:reward-models} lists the reward models used in our analysis.

\begin{table*}[ht]
\centering
\resizebox{.75\linewidth}{!}{
\begin{tabular}{llll}
    \toprule
    \textbf{Provider} & \textbf{Base Model} & \textbf{Size} & \textbf{Model ID}\\
    \midrule
    \multirow{5}{*}{Skywork} & \textsc{Llama-3.2} (Instruct) & 3B & \texttt{Skywork/Skywork-Reward-V2-Llama-3.2-3B}\\
    & \textsc{Llama-3.1} (Instruct) & 8B & \texttt{Skywork/Skywork-Reward-V2-Llama-3.1-8B}\\
    & \textsc{Qwen-3} (Instruct) & 4B & \texttt{Skywork/Skywork-Reward-V2-Qwen3-4B}\\
    & \textsc{Qwen-3} (Instruct) & 8B & \texttt{Skywork/Skywork-Reward-V2-Qwen3-8B}\\
    & \textsc{Gemma-2} (Instruct) & 27B & \texttt{Skywork/Skywork-Reward-Gemma-2-27B}\\
    \midrule
    \multirow{2}{*}{QRM} & \textsc{Llama-3.1} (Instruct) & 8B & \texttt{nicolinho/QRM-Llama3.1-8B-v2}\\
    & \textsc{Gemma-2} (Instruct) & 27B & \texttt{nicolinho/QRM-Gemma-2-27B}\\
    \midrule
    \multirow{3}{*}{Ai2} & \textsc{Llama-3.1} (Base) & 8B & \texttt{allenai/Llama-3.1-8B-Base-RM-RB2}\\
    & \textsc{Llama-3.1} (Instruct) & 8B & \texttt{allenai/Llama-3.1-8B-Instruct-RM-RB2}\\
    & \textsc{Llama-3.1} (Instruct) & 70B & \texttt{allenai/Llama-3.1-70B-Instruct-RM-RB2}\\
    \bottomrule
\end{tabular}
}
\caption{\textbf{Reward models used in our analysis.}}
\label{table:reward-models}
\end{table*}

\subsection{Computational Resources}

We ran our experiments on one shared compute server with 8 NVIDIA L4 GPUs and one shared compute server with 8 NVIDIA A5000 GPUs.

\section{Semantic Equivalence}
%%%%%%%%%%%%%%%%%%%%%%%%
% SEMANTIC EQUIVALENCE %
%%%%%%%%%%%%%%%%%%%%%%%%

\subsection{Perturbations}

\noindent\textbf{Swap.} We randomly replace each ASCII character with another random ASCII character. In our experiments, we swap characters with $\mathbb{P}(\text{swap}) = 0.05$.

\vspace{.5\baselineskip}

\noindent\textbf{Drop.} We randomly drop each character with a prespecified probability. In our experiments, we drop characters with $\mathbb{P}(\text{drop}) = 0.15$.

\vspace{.5\baselineskip}

\noindent\textbf{Insert.} We randomly insert an ASCII character with a prespecified probability. In our experiments, we insert characters with $\mathbb{P}(\text{insert}) = 0.05$.

\vspace{.5\baselineskip}

\noindent\textbf{Capitalize.} We have two methods of capitalizing our text: random and alternating. In the random case, we capitalize each character with a prespecified probability. In our experiments, we capitalize characters with $\mathbb{P}(\text{capitalize}) = 0.5$. In the alternating case, we alternate every other character.

\subsection{Translations}

We translate texts using the Google Translate API\footnote{\url{https://cloud.google.com/translate}}. Table~\ref{table:translation_transformation_details} lists the six languages we use, along with their corresponding API language codes.

\begin{table}[ht]
\centering
\resizebox{\linewidth}{!}{
\begin{tabular}{lcc}
\toprule
\textbf{Resource Level} & \textbf{Language} & \textbf{Google Translate Code} \\
\midrule
\multirow{2}{*}{High}  & Chinese & \texttt{zh-CN} \\
                        & French  & \texttt{fr} \\
\midrule
\multirow{2}{*}{Mid}   & Hindi   & \texttt{hi} \\
                        & Polish  & \texttt{pl} \\
\midrule
\multirow{2}{*}{Low}   & Khmer   & \texttt{km} \\
                        & Yoruba  & \texttt{yo} \\
\bottomrule
\end{tabular}
}
\caption{\textbf{Translation transformation details.}}
\label{table:translation_transformation_details}
\end{table}

\subsection{Perplexities}

We list input perplexities in Table~\ref{table:perturbation_perplexity_ratios}.

\begin{table}[ht]
\centering
\resizebox{\linewidth}{!}{
\begin{tabular}{lrrrrrr}
\toprule
& \multicolumn{2}{c}{\textbf{\textsc{MultiVALUE}}} & \multicolumn{2}{c}{\textbf{\textsc{ParallelAAVE}}} & \multicolumn{2}{c}{\textbf{\textsc{ReDial}}} \\
\cmidrule(lr){2-3} \cmidrule(lr){4-5} \cmidrule(lr){6-7}
\textbf{Condition} & \textbf{Ratio} & \textbf{frac$>$1} & \textbf{Ratio} & \textbf{frac$>$1} & \textbf{Ratio} & \textbf{frac$>$1} \\
\midrule
Drop ($\mathbb{P}=0.15$)          & $12.53$ & $1.00$ & $9.64$ & $1.00$ & $2.06$ & $1.00$ \\
\textit{Singapore}                & $4.92$  & $1.00$ & --     & --     & --     & --     \\
Insert ($\mathbb{P}=0.05$)        & $4.41$  & $1.00$ & $3.74$ & $0.98$ & $1.58$ & $1.00$ \\
\textit{Indian}                   & $3.99$  & $1.00$ & --     & --     & --     & --     \\
Drop ($\mathbb{P}=0.05$)          & $3.52$  & $1.00$ & $3.06$ & $0.98$ & $1.43$ & $1.00$ \\
\textit{AAVE}                     & $3.13$  & $1.00$ & $2.89$ & $0.92$ & $1.65$ & $0.96$ \\
Swap ($\mathbb{P}=0.05$)          & $3.08$  & $1.00$ & $2.33$ & $0.94$ & $1.36$ & $1.00$ \\
Capitalize (random)               & $2.36$  & $1.00$ & $1.46$ & $0.73$ & $1.41$ & $0.99$ \\
\textit{Appalachian}              & $2.10$  & $1.00$ & --     & --     & --     & --     \\
\textit{Chicano}                  & $1.46$  & $1.00$ & --     & --     & --     & --     \\
Translate (Yoruba)                & $1.38$  & $0.75$ & $0.68$ & $0.36$ & $1.26$ & $0.70$ \\
Translate (Chinese)               & $1.26$  & $0.80$ & $1.03$ & $0.50$ & $1.08$ & $0.74$ \\
Capitalize (alternating)          & $0.73$  & $0.37$ & $0.38$ & $0.14$ & $0.92$ & $0.40$ \\
Translate (Polish)                & $0.71$  & $0.16$ & $0.37$ & $0.10$ & $0.96$ & $0.39$ \\
Translate (French)                & $0.64$  & $0.08$ & $0.34$ & $0.05$ & $0.96$ & $0.34$ \\
Translate (Hindi)                 & $0.42$  & $0.21$ & $0.13$ & $0.10$ & $0.68$ & $0.23$ \\
Translate (Khmer)                 & $0.24$  & $0.15$ & $0.05$ & $0.11$ & $0.38$ & $0.17$ \\
\bottomrule
\end{tabular}
}
\caption{\textbf{Paired per-token perplexity ratio vs SAE.} We compare the perplexity ratios for character perturbations, translations, and dialects (\textit{italicized}). Each row pairs the transformed text to its matched SAE text by \texttt{unique\_id} within a model (ratio $= \exp(\text{CE}_{\text{cond}} - \text{CE}_{\text{sae}})$), aggregated as the median of per-model medians across the models listed in Table~\ref{table:language-models}. The mean fraction of paired items the model finds harder than SAE is indicated under \enquote{frac$>$1}, where a ratio above $1$ means the model is more surprised by the surface form than by SAE despite preserved meaning.}
\label{table:perturbation_perplexity_ratios}
\end{table}

\subsection{Results on \textsc{MultiVALUE}}

\begin{figure}[b]
\centering
\begin{subfigure}{\linewidth}
    \centering
    \includegraphics[width=\linewidth]{figures/multivalue_transformation_similarity.pdf}
    \caption{Semantic equivalence of transformations}
    \label{figure:semantic_equivalence_surface_difference-multivalue_taxes}
\end{subfigure}
\hfill
\begin{subfigure}{\linewidth}
    \centering
    \includegraphics[width=\linewidth]{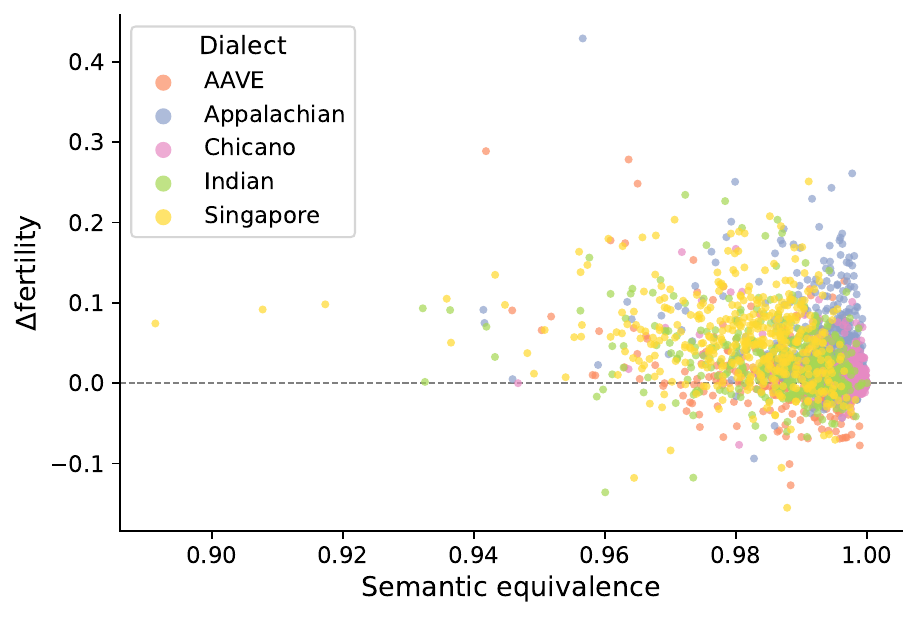}
    \caption{Semantic equivalence to $\Delta \text{fertility}$}
    \label{figure:semantic_equivalence_surface_difference-multivalue_similarities}
\end{subfigure}
\caption{\textbf{Models understand semantic equivalence yet penalize surface form.} We find evidence of semantic invariance under surface-form transformations, as shown on the \textsc{MultiVALUE} corpus. (\subref{figure:semantic_equivalence_surface_difference-multivalue_taxes}) We visualize the semantic equivalence of various text transformations. All dialect pairs achieve high cosine similarities above $0.97$, which exceed every perturbation and translation baseline. (\subref{figure:semantic_equivalence_surface_difference-multivalue_similarities}) We plot semantic equivalence (cosine similarity at $d = 768$) against the tokenization tax ($\Delta \text{fertility} = \mathbb{E}[\text{fertility}_{\text{dialect}}] - \mathbb{E}[\text{fertility}_{\text{SAE}}]$) for each dialect-SAE pair. Samples in the upper-right exhibit high semantic equivalence but higher tokenization cost, where meaning is preserved while a tax is imposed.}
\label{figure:semantic_equivalence_surface_difference_appendix}
\end{figure}

Figure~\ref{figure:semantic_equivalence_surface_difference-multivalue_taxes} visualizes Figure~\ref{figure:semantic_equivalence_surface_difference} from the main text and Figure~\ref{figure:semantic_equivalence_surface_difference-multivalue_similarities} visualizes the corresponding semantic equivalence to $\Delta \text{fertility}$.

\subsection{Results on \textsc{ParallelAAVE}}

\begin{figure}[b]
\centering
\begin{subfigure}{\linewidth}
    \centering
    \includegraphics[width=\linewidth]{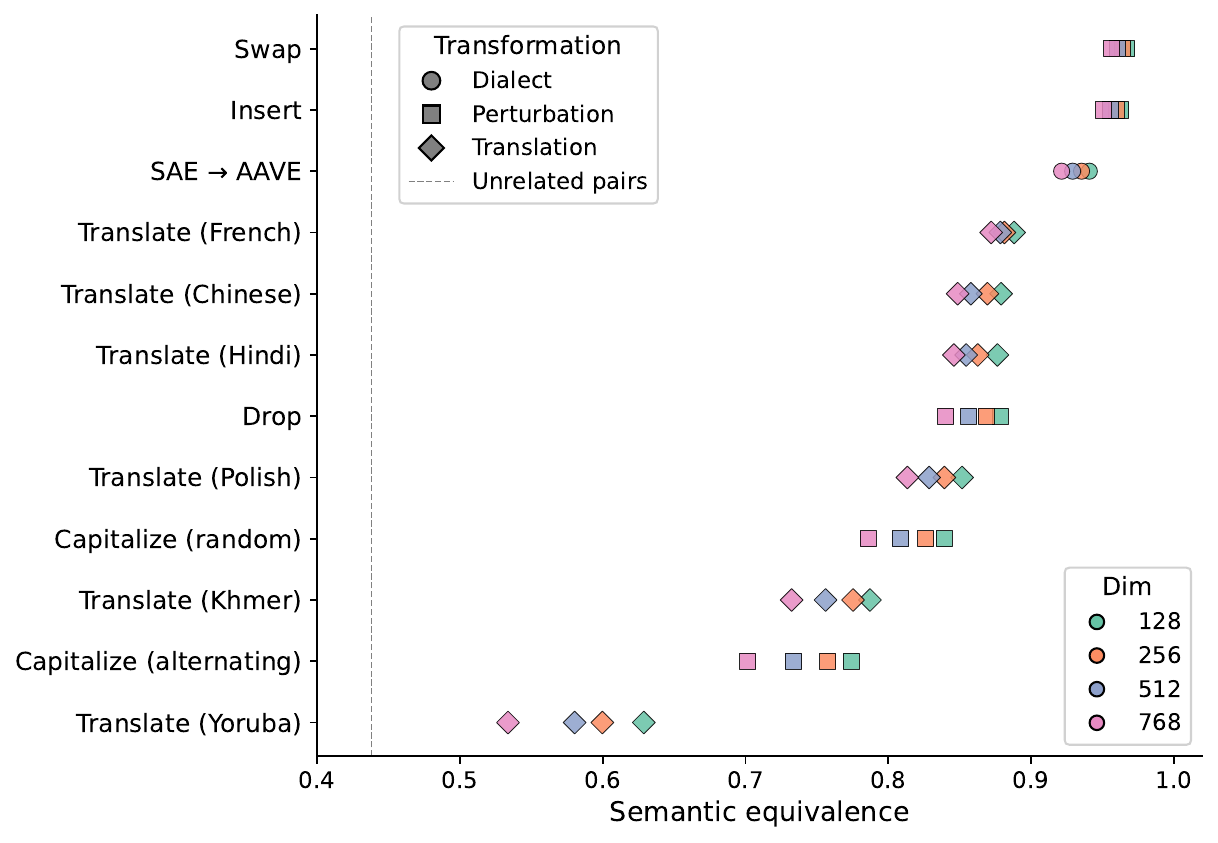}
    \caption{Semantic equivalence of transformations}
    \label{figure:semantic_equivalence_surface_difference-parallelaave_taxes}
\end{subfigure}
\hfill
\begin{subfigure}{\linewidth}
    \centering
    \includegraphics[width=\linewidth]{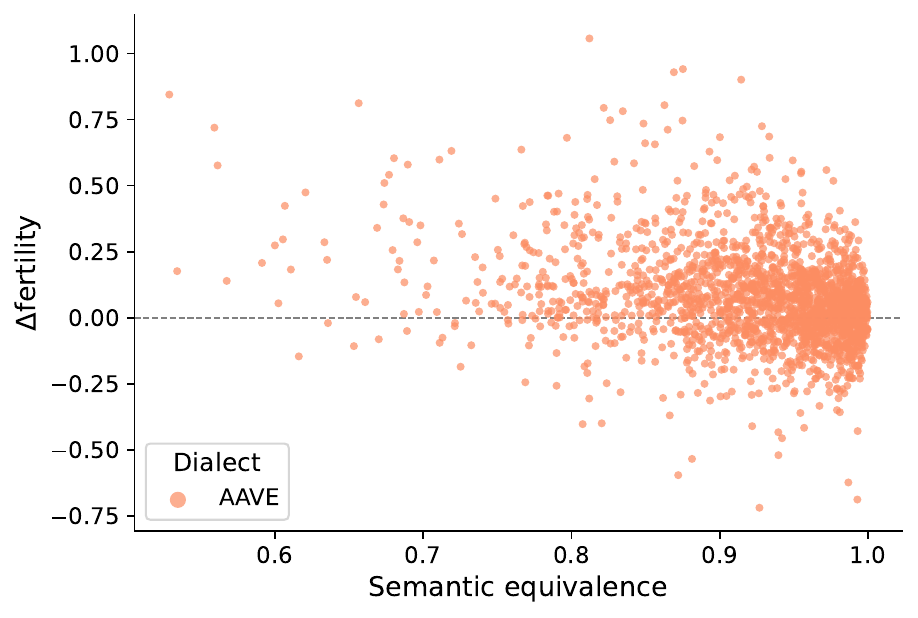}
    \caption{Semantic equivalence to $\Delta \text{fertility}$}
    \label{figure:semantic_equivalence_surface_difference-parallelaave_similarities}
\end{subfigure}
\caption{\textbf{Models understand semantic equivalence yet penalize surface form.} We find evidence of semantic invariance under surface-form transformations, as shown on the \textsc{ParallelAAVE} corpus. (\subref{figure:semantic_equivalence_surface_difference-parallelaave_taxes}) We visualize the semantic equivalence of various text transformations. The AAVE pairs achieve high cosine similarities, which exceed most perturbation and translation baselines. (\subref{figure:semantic_equivalence_surface_difference-parallelaave_similarities}) We plot semantic equivalence (cosine similarity at $d = 768$) against the tokenization tax ($\Delta \text{fertility} = \mathbb{E}[\text{fertility}_{\text{AAVE}}] - \mathbb{E}[\text{fertility}_{\text{SAE}}]$) for each AAVE-SAE pair. Samples in the upper-right exhibit high semantic equivalence but higher tokenization cost, where meaning is preserved while a tax is imposed.}
\label{figure:semantic_equivalence_surface_difference_parallelaave}
\end{figure}

Figure~\ref{figure:semantic_equivalence_surface_difference_parallelaave} visualizes the same figure as Figure~\ref{figure:semantic_equivalence_surface_difference_appendix} on the \textsc{ParallelAAVE} corpus.

\section{Benchmarks}
%%%%%%%%%%%%%%
% BENCHMARKS %
%%%%%%%%%%%%%%

\subsection{Tokenization Biases}
\label{appendix:tokenization-biases}

\noindent\textbf{Metrics.} Table~\ref{table:tokenization-bias-metrics} lists all token metrics used in our study. We define a \enquote{token} as an element of a tokenizer vocabulary. We define \enquote{types} as the set of unique tokens. We define a \enquote{word} (in English) as a string without any space characters. We define a \enquote{character} as any string with a length of $1$.

\begin{table}[t]
\centering
\resizebox{\linewidth}{!}{
\begin{tabular}{p{0.3\linewidth} p{0.7\linewidth}}
    \toprule
    \textbf{Metric} & \textbf{What does it measure?}\\
    \midrule
    Average tokens per word & {Average number of tokens corresponding with a single real word without punctuations}\\
    \midrule
    Average types per word & {Average number of types corresponding with a single real word without punctuations}\\
    \midrule
    Character length & {Number of characters in the string}\\
    \midrule
    Fertility \citep{fertility} & {Average number of tokens corresponding with a single real word}\\
    \midrule
    P(in vocabulary) & {Proportion of words in the tokenizer vocabulary}\\
    \midrule
    Token length & {Number of tokens in the string}\\
    \midrule
    Types length & {Number of types in the string}\\
    \midrule
    Word length & {Number of words in the string}\\
    \bottomrule
\end{tabular}}
\caption{\textbf{Tokenization metrics used to measure bias.}}
\label{table:tokenization-bias-metrics}
\end{table}

\vspace{.5\baselineskip}

\begin{figure*}[ht]
\centering
\includegraphics[width=\textwidth]{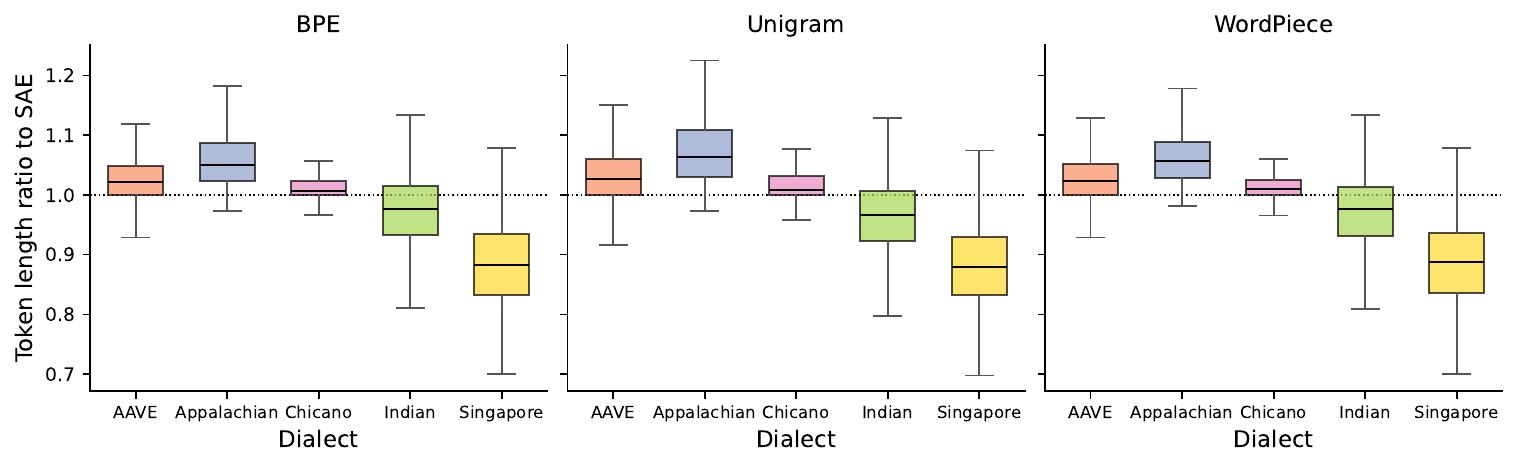}
\caption{\textbf{Ratio of dialect to SAE token lengths on \textsc{MultiVALUE}.} Dotted lines indicate token-length parity with the paired SAE text. We see similar tokenization bias ranking on every measured metric for the six dialects in the dataset (Appalachian $>$ AAVE $>$ Chicano $>$ SAE $>$ Indian $>$ Singapore). While some dialects (AAVE, Appalachian, Chicano) have increased token lengths compared to SAE, other dialects (Indian, Singapore) have decreased token lengths compared to SAE. We find that the same dialectal token bias remains present across all three tokenization strategies.}
\label{figure:tokenizer-dialect-bias-ratio-multivalue}
\end{figure*}

\noindent\textbf{Token Length.} Figure~\ref{figure:tokenizer-dialect-bias-ratio-multivalue} visualizes the token length ratio of various dialects to SAE on \textsc{MultiVALUE} for all three tokenization algorithms.

\vspace{.5\baselineskip}

\begin{figure}[H]
\centering
\includegraphics[width=\linewidth]{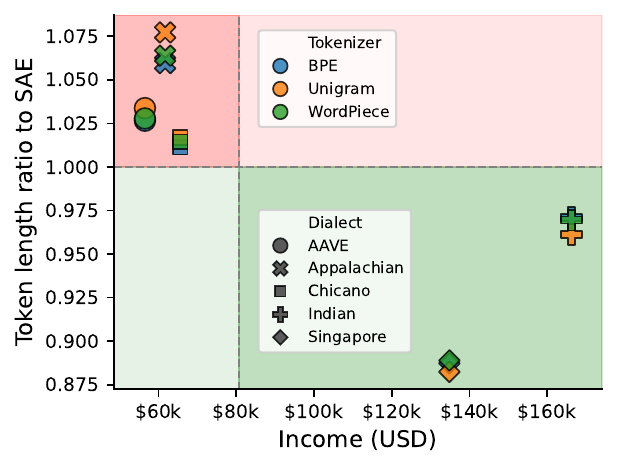}
\caption{\textbf{Dialectal token bias and income.} We demarcate token parity with $y = 1$ and the US median household income with $x = \$80,610$. Compared to Standard American English (SAE) speakers, speakers of African American Vernacular English (AAVE), Appalachian, and Chicano dialects who earn less than the median income achieve less efficient tokenization. Meanwhile, speakers of Indian and Singaporean dialects who earn more than the median income achieve more efficient tokenization than SAE speakers.}
\label{figure:tokenization-bias-and-income}
\end{figure}

\noindent\textbf{Token bias and income.}  Figure~\ref{figure:tokenization-bias-and-income} depicts the striking correspondence between dialectal token length ratios to SAE and the current income differences of minority groups in the US. This tokenization gap mirrors the current income gaps of minority groups within the US. Speakers of AAVE and Chicano dialects are typically black or Hispanic, respectively, who report the lowest median household incomes among all race and ethnic groups \citep{medianincomeminorityusa}. Median household income within the Appalachian Region is $82$\% of that of America overall \citep{medianincomeappalachiausa}. Meanwhile, Indian Americans on average see much higher incomes than the total American population \citep{medianincomeindiansusa}, and Singapore is among the wealthiest countries in the world, with a GDP per capita surpassing that of America \citep{gdppercapitasingapore}.

Table~\ref{table:median-household-income} details the US median household incomes (in US dollars) for different dialects. We approximate the median household income per dialect by using the median household income data of the primary demographic of the speakers of that dialect. That is, we take the median income of Black Americans to approximate the median income of AAVE speakers, and we take the median income of Hispanic Americans to approximate the median income of Chicano speakers. We use the overall median household income of the US to approximate the median income of SAE speakers.

\begin{table}[H]
\centering
\resizebox{\linewidth}{!}{\begin{tabular}{p{0.23\linewidth} p{0.18\linewidth} p{0.55\linewidth}}
    \toprule
    \textbf{Dialect} & \textbf{Income} & \textbf{Source}\\
    \midrule
    AAVE & $\$56,490$ & \citet{medianincomeminorityusa}\\
    \midrule
    Appalachian & $\$61,688$ & \citet{medianincomeappalachiausa}\\
    \midrule
    Chicano & $\$65,540$ & \citet{medianincomeminorityusa}\\
    \midrule
    Indian & $\$166,200$ & \citet{medianincomeindiansusa}\\
    \midrule
    SAE &  $\$80,610$ & \citet{medianincomeminorityusa}\\
    \midrule
    Singaporean & $\$134,818$ & \citet{medianincomesingaporeansusa}\\
    \bottomrule
\end{tabular}}
\caption{\textbf{US median household income (USD).}}
\label{table:median-household-income}
\end{table}

\vspace{.5\baselineskip}

\begin{figure*}[ht]
\centering
\includegraphics[width=\textwidth]{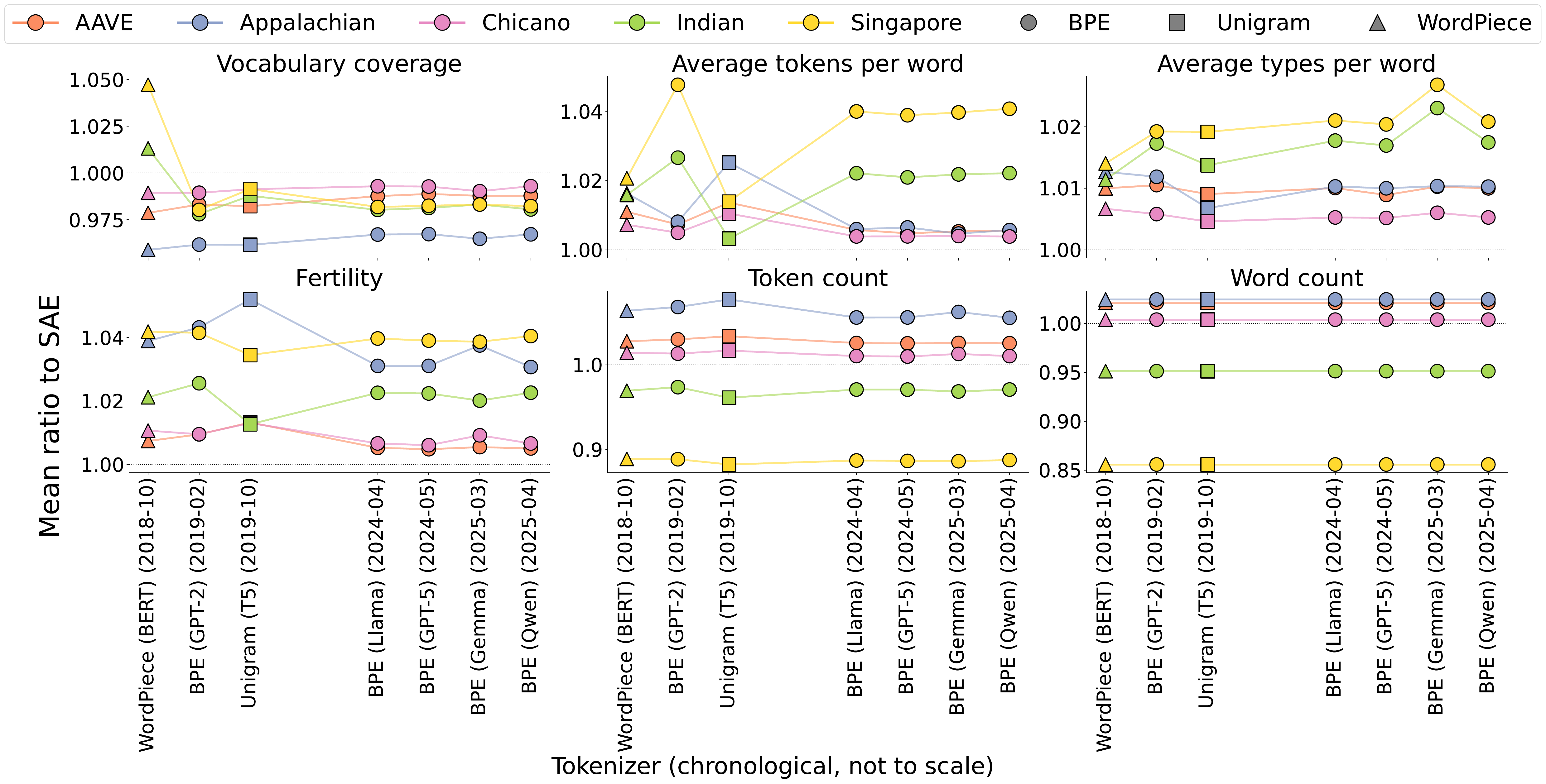}
\caption{\textbf{Tokenizer metric timeline.} We plot a subset of the token metrics in Table~\ref{table:tokenization-bias-metrics} over time.}
\label{figure:tokenizer-metric-timeline}
\end{figure*}

\noindent\textbf{Token metric timeline.} Figure~\ref{figure:tokenizer-metric-timeline} shows the chronological development of our token metrics. While we find an overall improvement in vocabulary coverage cover time, we still see evidence of the dialect tax in today's tokenizers. To quantify ranking stability, we compute Kendall's coefficient of concordance $W$ across the seven tokenizers, treating each tokenizer as a rater of the five non-SAE dialects. Agreement is statistically significantly ($p < 0.001$) strong on every metric (fertility $W = 0.89$, $\chi^2(4) = 24.80$; vocabulary coverage $W = 0.74$; average tokens per word $W = 0.77$; average types per word $W = 0.95$; token count and word count both $W = 1.00$), indicating that the dialectal ordering is essentially fixed across tokenizers. This concordance holds despite seven years of tokenizer development. For each dialect, we correlate the tokenizer's release date with its rank among the five dialects under that tokenizer (Spearman $\rho$, Benjamini-Hochberg FDR at $q = 0.05$). Zero of the $30$ metric $\times$ dialect cells show significant rank drift, and every dialect holds a constant rank across all seven tokenizers on token count and word count. Complementary tests on the magnitude of the gap also fail to survive correction. Spearman correlations between the release date and the mean log-ratio to SAE are largest on fertility for AAVE, Appalachian, and Chicano ($\rho \approx -0.75$, $p_{\mathrm{raw}} \in [0.02, 0.07]$), consistent with a mild improvement for those dialects, but none of the $30$ comparisons survive BH correction.

\subsection{Reasoning}
\label{appendix:benchmark_reasoning}

\begin{table*}[t]
\centering
\makebox[\textwidth][c]{
\resizebox{1\textwidth}{!}{
\begin{tabular}{llccccccccccc}
\toprule
\multirow{2}{*}{\textbf{Model}} & \multirow{2}{*}{\textbf{Setting}} & \multicolumn{2}{c}{\textbf{Algorithm}} & \multicolumn{2}{c}{\textbf{Logic}} & \multicolumn{2}{c}{\textbf{Math}} & \multicolumn{2}{c}{\textbf{Planning}} & \multicolumn{3}{c}{\textbf{All}} \\
 &  & SAE & AAVE & SAE & AAVE & SAE & AAVE & SAE & AAVE & SAE & AAVE & $\Delta$ \\
\midrule
\multirow[t]{2}{*}{\textsc{GPT}} & CoT & 93.6 & 90.4 & 80.4 & 76.8 & 94.0 & 92.0 & 96.4 & 92.9 & 91.1 & 88.0 & \textbf{-3.09} \\
 & Na\"ive & 91.7 & 89.1 & 68.0 & 66.3 & 76.3 & 69.3 & 17.3 & 13.8 & 63.3 & 59.6 & \textbf{-3.69} \\
\noalign{\vskip\aboverulesep}\cdashline{2-13}\noalign{\vskip\belowrulesep}
\multirow[t]{2}{*}{\textsc{GPT} Mini} & CoT & 93.0 & 90.4 & 77.1 & 69.9 & 92.7 & 90.0 & 97.8 & 96.4 & 90.1 & 86.7 & -3.43 \\
 & Na\"ive & 91.4 & 88.2 & 66.9 & 59.9 & 64.0 & 54.0 & 6.7 & 10.7 & 57.2 & 53.2 & -4.03 \\
\midrule
\multirow[t]{2}{*}{\textsc{Gemma 27B}} & CoT & 83.7 & 80.5 & 66.0 & 66.9 & 92.7 & 89.0 & 24.4 & 24.0 & 66.7 & 65.1 & \textbf{-1.62} \\
 & Na\"ive & 82.7 & 79.6 & 66.6 & 65.2 & 54.3 & 50.7 & 1.3 & 1.3 & 51.2 & 49.2 & \textbf{-2.06} \\
\noalign{\vskip\aboverulesep}\cdashline{2-13}\noalign{\vskip\belowrulesep}
\multirow[t]{2}{*}{\textsc{Gemma 12B}} & CoT & 84.0 & 77.3 & 76.2 & 71.5 & 92.3 & 87.7 & 3.1 & 2.2 & 63.9 & 59.7 & \textbf{-4.24} \\
 & Na\"ive & 79.9 & 75.7 & 72.4 & 69.6 & 46.3 & 39.7 & 0.0 & 0.4 & 49.6 & 46.4 & \textbf{-3.28} \\
\midrule
\multirow[t]{2}{*}{\textsc{Llama 70B}} & CoT & 81.2 & 81.2 & 74.3 & 69.9 & 91.3 & 89.3 & 48.0 & 36.9 & 73.7 & 69.3 & -4.38 \\
 & Na\"ive & 83.1 & 79.9 & 74.6 & 73.8 & 57.3 & 49.7 & 0.9 & 0.9 & 54.0 & 51.0 & -2.92 \\
\noalign{\vskip\aboverulesep}\cdashline{2-13}\noalign{\vskip\belowrulesep}
\multirow[t]{2}{*}{\textsc{Llama 8B}} & CoT & 62.0 & 55.9 & 63.5 & 63.8 & 88.3 & 77.0 & 9.8 & 4.0 & 55.9 & 50.2 & \textbf{-5.73} \\
 & Na\"ive & 58.5 & 55.9 & 66.0 & 58.0 & 41.3 & 31.0 & 0.0 & 0.0 & 41.5 & 36.2 & \textbf{-5.23} \\
\midrule
\multirow[t]{2}{*}{\textsc{Qwen 27B}} & CoT & 95.2 & 93.6 & 79.3 & 73.2 & 96.0 & 92.3 & 98.7 & 92.4 & 92.3 & 87.9 & \textbf{-4.39} \\
 & Na\"ive & 55.0 & 54.0 & 73.5 & 67.1 & 70.7 & 68.7 & 2.2 & 0.9 & 50.3 & 47.7 & -2.66 \\
\noalign{\vskip\aboverulesep}\cdashline{2-13}\noalign{\vskip\belowrulesep}
\multirow[t]{2}{*}{\textsc{Qwen 8B}} & CoT & 91.7 & 89.1 & 67.4 & 61.9 & 96.3 & 92.7 & 68.9 & 74.7 & 81.1 & 79.6 & -1.49 \\
 & Na\"ive & 69.0 & 65.5 & 47.8 & 43.9 & 34.7 & 33.0 & 0.9 & 0.4 & 38.1 & 35.7 & -2.37 \\
\bottomrule
\end{tabular}
}}
\caption{\textbf{Benchmark results on \textsc{ReDial}.} We report instruction-tuned model accuracies on \textsc{ReDial}, including individual performances across the four tasks and the overall performance under the column \enquote{All}. In accordance with the original \textsc{ReDial} benchmark, we test for statistical significance (indicated in bold under the column \enquote{$\Delta$}) using McNemar's test for binary data and correct the $p$-values for multiple measurements using the Holm-Bonferroni method.}
\label{table:benchmark-redial}
\end{table*}

Table~\ref{table:benchmark-redial} shows the benchmark results for \textsc{ReDial}. The values track with the expected progress of LMs since the original dataset publication by \citet{redial}, given the modifications specified in Appendix~\ref{appendix:redial}. We note that \textsc{Qwen} models always prefer to reason via \texttt{<think>...</think>} tokens, even on na\"ive prompts explicitly asking for no reasoning, so we append a model-family-specific prompt to ensure no reasoning.

\section{Character Tokenization}
%%%%%%%%%%%%%%%%%%%%%%%%%%
% CHARACTER TOKENIZATION %
%%%%%%%%%%%%%%%%%%%%%%%%%%
\label{appendix:character-tokenization}

The character-tokenization experiments replace subword segmentation with a non-canonical character decomposition at inference time. This intervention keeps the model vocabulary, embedding matrix, tokenizer, and learned parameters fixed. We follow the method by \citet{zheng2026brokentokenslanguagemodel} exactly.

\section{Pre-Training Analysis}
\label{appendix:pretraining-analysis}
%%%%%%%%%%%%%%%%
% PRE-TRAINING %
%%%%%%%%%%%%%%%%

Table~\ref{table:gradient_similarity_correctness_correlation} lists the point-biserial correlation between $s_i^+$ and correctness by base model.

\begin{table}[ht]
\centering
\resizebox{.7\linewidth}{!}{
\begin{tabular}{llr}
\toprule
\textbf{Family} & \textbf{Size} & $r$ \\
\midrule
\multirow{3}{*}{\textsc{Llama-3 Base}}
                                & 1B  & $-0.119^{***}$ \\
                                & 3B  & $-0.053$      \\
                                & 8B  & $-0.091^{**}$  \\
\midrule
\multirow{3}{*}{\textsc{Gemma-3 Base}}
                                & 1B  & $-0.108^{***}$ \\
                                & 4B  & $-0.105^{***}$ \\
                                & 12B & $-0.066$      \\
\midrule
\multirow{3}{*}{\textsc{Qwen-3 Base}}
                                & 1.7B & $-0.132^{***}$ \\
                                & 4B   & $+0.098^{***}$ \\
                                & 8B   & $+0.192^{***}$ \\
\midrule
\textbf{Pooled}                  & --   & $-0.013$      \\
\bottomrule
\end{tabular}
}
\caption{\textbf{Point-biserial correlation between paired SAE-AAVE gradient similarity $s_i^+$ and a binary indicator for both-dialect correctness, by model.} Per-model sample size is $n = 1{,}200$, and pooled sample size is $n = 10{,}800$. We denote significance by ${}^{***}\,p < 0.001$ and ${}^{**}\,p < 0.01$. Pooled $r$ is negligible, but the per-model breakdown reveals heterogeneity.}
\label{table:gradient_similarity_correctness_correlation}
\end{table}

\section{Post-Training Analysis}
\label{appendix:posttraining-analysis}
%%%%%%%%%%%%%%%%%
% POST-TRAINING %
%%%%%%%%%%%%%%%%%

We list all reward models used in Table~\ref{table:reward-models}.

\subsection{Sample-Level Rewards}

Table~\ref{table:rewards_null_test_by_rm} lists the per-RM dialect gap for $\overline{\Delta r}$. Table~\ref{table:rewards_null_test_by_task} denotes the sample-level rewards per task, pooled across RMs.

\begin{table}[ht]
\centering
\small
\begin{tabular}{lrr}
\toprule
\textbf{Reward model} & $\overline{\Delta r}$ & $t$ \\
\midrule
\textsc{Skywork}~\textsc{Llama} 3B   & $-0.03$ & $-0.67$         \\
\textsc{Skywork}~\textsc{Qwen} 4B    & $+0.32$ & $+7.54^{***}$   \\
\textsc{Skywork}~\textsc{Llama} 8B   & $-0.22$ & $-3.27^{**}$    \\
\textsc{Skywork}~\textsc{Qwen} 8B    & $+0.48$ & $+14.42^{***}$  \\
\textsc{Skywork}~\textsc{Gemma} 27B  & $-0.03$ & $-0.31$         \\
\midrule
\textsc{QRM}~\textsc{Llama} 8B       & $+0.01$ & $+4.72^{***}$   \\
\textsc{QRM}~\textsc{Gemma} 27B      & $-0.07$ & $-3.38^{***}$   \\
\midrule
\textsc{Ai2}~\textsc{Llama} 8B Base  & $+0.41$ & $+12.28^{***}$  \\
\textsc{Ai2}~\textsc{Llama} 8B       & $-0.19$ & $-10.75^{***}$  \\
\textsc{Ai2}~\textsc{Llama} 70B      & $+0.03$ & $+2.06^{*}$     \\
\bottomrule
\end{tabular}
\caption{\textbf{Per-RM dialect gap.} $\overline{\Delta r} = \mathbb{E}[r(x^{\text{SAE}},y) - r(x^{\text{AAVE}},y)]$ over \textsc{ReDial} sample-level pairs ($n = 1{,}200$ per RM, i.e. 300 per
task $\times$ 4 tasks), and one-sample $t$-statistic against $\mathbb{E}[\Delta r] = 0$. Significance is denoted by ${}^{***}\,p < 0.001$, ${}^{**}\,p < 0.01$, ${}^{*}\,p < 0.05$.}
\label{table:rewards_null_test_by_rm}
\end{table}

\begin{table}[ht]
\centering
\small
\begin{tabular}{lrr}
\toprule
\textbf{Task} & $\overline{\Delta r}$ & $t$ \\
\midrule
Algorithm & $+0.43$ & $+13.73^{***}$ \\
Math      & $+0.04$ & $+1.18$        \\
Logic     & $-0.08$ & $-2.92^{**}$   \\
Planning  & $-0.14$ & $-4.92^{***}$  \\
\bottomrule
\end{tabular}
\caption{\textbf{Per-task dialect gap, pooled across RMs.} One-sample $t$-test of $\mathbb{E}[\Delta r] = 0$ within task. Significance is denoted by ${}^{***}\,p < 0.001$, ${}^{**}\,p < 0.01$.}
\label{table:rewards_null_test_by_task}
\end{table}

\begin{table*}[ht]
\centering
\resizebox{\linewidth}{!}{
\begin{tabular}{lcrrrrrrr}
\toprule
\textbf{Corpus} & \textbf{Dialects} & $n_{\text{SAE}}$ & $n_{\text{dialect}}$ & $\overline{r}_{\text{SAE}}$ & $\overline{r}_{\text{dialect}}$ & $\overline{r}_{\text{SAE}} - \overline{r}_{\text{dialect}}$ &
$\overline{d}_{\text{RM}}$ & $p$ \\
\midrule
\textsc{ReDial}       & \{AAVE\}                                                & 121{,}780 & 163{,}120 & $-4.47$ & $-4.10$ & $-0.37^{***}$ & $-0.17$ & $9 \times 10^{-31}$  \\
\textsc{ParallelAAVE} & \{AAVE\}                                                & 91{,}380  & 163{,}420 & $-4.42$ & $-3.83$ & $-0.58^{***}$ & $-0.29$ & $1 \times 10^{-67}$  \\
\textsc{MultiVALUE}   & \{AAVE, Appal., Chic., Indian, Sing.\}                  & 141{,}930 & 132{,}130 & $-5.17$ & $-4.74$ & $-0.43^{***}$ & $-0.24$ & $9 \times 10^{-39}$  \\
\midrule
\textbf{Pooled (all)} &                                                         & 355{,}090 & 458{,}670 & $-4.74$ & $-4.19$ & $-0.55^{***}$ & $-0.27$ & $7 \times 10^{-185}$ \\
\bottomrule
\end{tabular}}
\caption{\textbf{Per-corpus token-level dialect gap.} For each corpus, we identify subword tokens that appear exclusively in tokenized SAE vs. dialect text within each (RM, tokenizer) pairing, and score them under a fixed prompt. $\overline{r}_{\text{SAE}}$ and $\overline{r}_{\text{dial}}$ are the mean reward scores over the dialect-exclusive vocabularies. $\overline{r}_{\text{SAE}} - \overline{r}_{\text{dial}}$ is the raw gap. $\overline{d}_{\text{RM}}$ is Cohen's $d$ computed within each RM (using that RM's pooled score standard deviation) and then averaged across the ten RMs. This normalizes for the substantial cross-RM scale differences (per-RM $\sigma \in [0.06, 3.97]$). The bottom row pools all three corpora together. Independent two-sample $t$-tests, with significance denoted by ${}^{***}\,p < 0.001$.}
\label{table:rewards_token_by_corpus}
\end{table*}

\subsection{Token-Level Rewards}

Unique words are extracted from each dataset split by whitespace and delimiter-based word segmentation. Unique tokens are extracted under each tokenizer listed in Table~\ref{table:tokenizers}. Each word is scored as a one-word assistant response to the fixed prompt \enquote{\texttt{What, in one word, is the greatest thing ever?}}, and each token is scored as a one-word-or-subword assistant response to \enquote{\texttt{What, in one word or subword, is the greatest thing ever?}}. For token-level dialect comparisons, for the same task and tokenizer, we identify tokens that appear in AAVE but not SAE and tokens that appear in SAE but not AAVE, then compare their isolated reward scores. Table~\ref{table:rewards_token_per_rm_dialect} lists the per-RM per-dialect token-level dialect gap. Table~\ref{table:rewards_token_by_dialect_scaling} details the per-dialect token-level gap, before and after within-RM standardization.

\begin{table*}[ht]
\centering
\resizebox{\linewidth}{!}{
\begin{tabular}{lcccccc}
\toprule
\textbf{Reward model} & \textbf{AAVE} & \textbf{Appalachian} & \textbf{Chicano} & \textbf{Indian} & \textbf{Singaporean} & \textbf{Pooled} \\
\midrule
\textsc{Skywork}~\textsc{Llama}~3B    & $-0.45^{***}$ ($-0.25$) & $+0.11^{*}$ ($+0.06$)   & $-0.71^{***}$ ($-0.43$) & $-0.06$ ($-0.03$)        & $-0.88^{***}$ ($-0.49$) & $-0.51^{***}$ ($-0.28$) \\
\textsc{Skywork}~\textsc{Qwen}~4B     & $-0.69^{***}$ ($-0.31$) & $-0.32^{***}$ ($-0.15$) & $-0.71^{***}$ ($-0.37$) & $-0.13^{*}$ ($-0.06$)    & $-1.10^{***}$ ($-0.47$) & $-0.78^{***}$ ($-0.34$) \\
\textsc{Skywork}~\textsc{Llama}~8B    & $-1.26^{***}$ ($-0.38$) & $-0.09$ ($-0.03$)       & $-1.49^{***}$ ($-0.49$) & $-0.28^{**}$ ($-0.08$)   & $-1.34^{***}$ ($-0.38$) & $-1.26^{***}$ ($-0.37$) \\
\textsc{Skywork}~\textsc{Qwen}~8B     & $-0.80^{***}$ ($-0.40$) & $-0.16^{**}$ ($-0.09$)  & $-0.67^{***}$ ($-0.35$) & $+0.00$ ($+0.00$)        & $-0.61^{***}$ ($-0.29$) & $-0.73^{***}$ ($-0.36$) \\
\textsc{Skywork}~\textsc{Gemma}~27B   & $-1.04^{***}$ ($-0.27$) & $-1.74^{***}$ ($-0.45$) & $-1.97^{***}$ ($-0.54$) & $\mathbf{+2.15^{***}}$ $\mathbf{(+0.57)}$ & $-2.08^{***}$ ($-0.52$) & $-1.17^{***}$
($-0.30$) \\
\midrule
\textsc{QRM}~\textsc{Llama}~8B        & $+0.00^{***}$ ($+0.04$) & $-0.00$ ($-0.01$)       & $-0.02^{***}$ ($-0.37$) & $-0.00$ ($-0.05$)        & $-0.01^{***}$ ($-0.19$) & $-0.00^{*}$ ($-0.02$)   \\
\textsc{QRM}~\textsc{Gemma}~27B       & $-0.15^{***}$ ($-0.30$) & $-0.18^{***}$ ($-0.37$) & $-0.23^{***}$ ($-0.46$) & $-0.12^{***}$ ($-0.24$)  & $-0.17^{***}$ ($-0.34$) & $-0.16^{***}$ ($-0.33$) \\
\midrule
\textsc{Ai2}~\textsc{Llama}~8B Base   & $-0.42^{***}$ ($-0.26$) & $-0.39^{***}$ ($-0.25$) & $-1.05^{***}$ ($-0.74$) & $-0.38^{***}$ ($-0.22$)  & $-1.19^{***}$ ($-0.71$) & $-0.55^{***}$ ($-0.34$) \\
\textsc{Ai2}~\textsc{Llama}~8B        & $-0.09^{***}$ ($-0.12$) & $-0.07^{***}$ ($-0.10$) & $-0.28^{***}$ ($-0.47$) & $-0.07^{***}$ ($-0.09$)  & $-0.30^{***}$ ($-0.43$) & $-0.13^{***}$ ($-0.18$) \\
\textsc{Ai2}~\textsc{Llama}~70B       & $-0.16^{***}$ ($-0.15$) & $-0.02$ ($-0.03$)       & $-0.13^{**}$ ($-0.17$)  & $-0.05^{*}$ ($-0.05$)    & $-0.22^{***}$ ($-0.22$) & $-0.18^{***}$ ($-0.17$) \\
\midrule
\textbf{Mean across RMs}              & $-0.50$ ($-0.24$)       & $-0.29$ ($-0.14$)       & $-0.73$ ($-0.44$)       & $+0.11$ ($-0.03$)        & $-0.79$ ($-0.40$)       & $-0.55$ ($-0.27$)       \\
\bottomrule
\end{tabular}
}
\caption{\textbf{Per-RM per-dialect token-level dialect gap.} Each cell shows the raw reward gap $\overline{r}_{\text{SAE}} - \overline{r}_{\text{dial}}$ with Cohen's $d$ in parentheses. Negative values indicate the RM scores dialect-exclusive tokens higher than SAE-exclusive ones. AAVE-exclusive tokens come from \textsc{ReDial}, \textsc{ParallelAAVE}, and \textsc{MultiVALUE}, and other dialect-exclusive tokens come from \textsc{MultiVALUE}. Cohen's $d$ uses each RM's pooled score standard deviation as denominator, so raw gaps are not directly comparable across RMs while $d$ values are. The bolded cell is the only one flipping the dialect-favoring pattern. The ``Pooled'' column aggregates a given RM across all five dialects, and the ``Mean across RMs'' row is the unweighted average across the ten RMs. We run independent two-sample $t$-tests and denote the significance: ${}^{***}\,p < 0.001$, ${}^{**}\,p < 0.01$, ${}^{*}\,p < 0.05$.}
\label{table:rewards_token_per_rm_dialect}
\end{table*}

\begin{table*}[t]
\centering
\resizebox{.7\linewidth}{!}{
\begin{tabular}{lrrrrrr}
\toprule
& & & \multicolumn{2}{c}{\textbf{Raw scores}} & \multicolumn{2}{c}{\textbf{Within-RM standardized}} \\
\cmidrule(lr){4-5} \cmidrule(lr){6-7}
\textbf{Dialect} & $n_{\text{SAE}}$ & $n_{\text{dial}}$ & $\overline{r}_{\text{SAE}} - \overline{r}_{\text{dial}}$ & $p$ & $d$ & $p$ \\
\midrule
AAVE         & 239{,}000 & 351{,}140 & $-0.50$ & $6 \times 10^{-115}$ & $-0.24$ & $< 10^{-300}$ \\
Appalachian  &  24{,}980 &  31{,}540 & $-0.29$ & $8 \times 10^{-5}$   & $-0.13$ & $5 \times 10^{-64}$  \\
Chicano      &   8{,}960 &   6{,}640 & $-0.73$ & $3 \times 10^{-7}$   & $-0.39$ & $2 \times 10^{-165}$ \\
Indian       &  27{,}200 &  32{,}710 & $+0.11$ & $0.14$ & $-0.03$ & $4 \times 10^{-4}$ \\
Singaporean  &  54{,}950 &  36{,}640 & $-0.79$ & $6 \times 10^{-40}$  & $-0.40$ & $< 10^{-300}$ \\
\midrule
\textbf{All pooled} & 355{,}090 & 458{,}670 & $-0.55$ & $7 \times 10^{-185}$ & $-0.27$ & $< 10^{-300}$ \\
\bottomrule
\end{tabular}
}
\caption{\textbf{Per-dialect token-level gap, before and after within-RM standardization.} \textit{Raw scores:} pooled two-sample $t$-test on raw reward scores. Note that per-RM output scales differ, so the test is heavily influenced by the high-$\sigma$ Skywork models. \textit{Within-RM standardized:} each RM's scores are $z$-scored using its own mean and standard deviation, then pooled across RMs before the $t$-test. The $d$ column reports the gap in standard deviations, which is mathematically equivalent to the unweighted mean Cohen's $d$ across the ten RMs (within-RM effect size).}
\label{table:rewards_token_by_dialect_scaling}
\end{table*}

\section{Inference}
\label{appendix:inference}
%%%%%%%%%%%%%
% INFERENCE %
%%%%%%%%%%%%%

\begin{figure}[H]
\centering
\includegraphics[width=\linewidth]{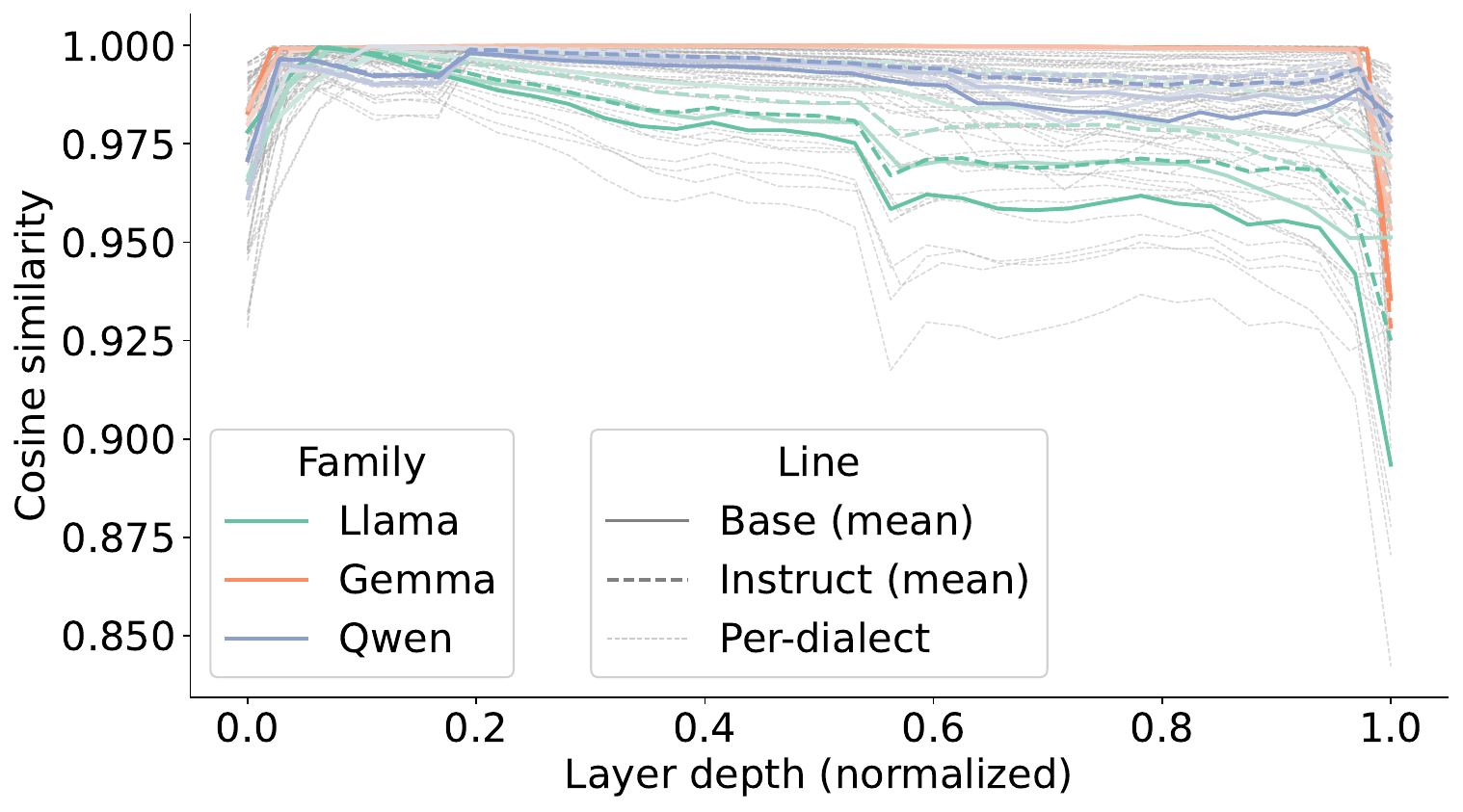}
\caption{\textbf{Hidden-state similarity drops at the final layers for every model on \textsc{MultiVALUE}.} Per-model layer-wise cosine similarity between SAE and dialect hidden states on \textsc{MultiVALUE}. Grey dashed curves show each of the five dialects (AAVE, Appalachian, Chicano, Indian, Singaporean) separately; colored curves are the per-model mean across the five dialects (solid = base, dashed = instruct), with color encoding family.}
\label{figure:multivalue_aggregate_all_models}
\end{figure}

Figure~\ref{figure:multivalue_aggregate_all_models} visualizes the hidden-state similarity dialect drops for every model on \textsc{MultiVALUE}.

\begin{figure}[H]
\centering
\includegraphics[width=\linewidth]{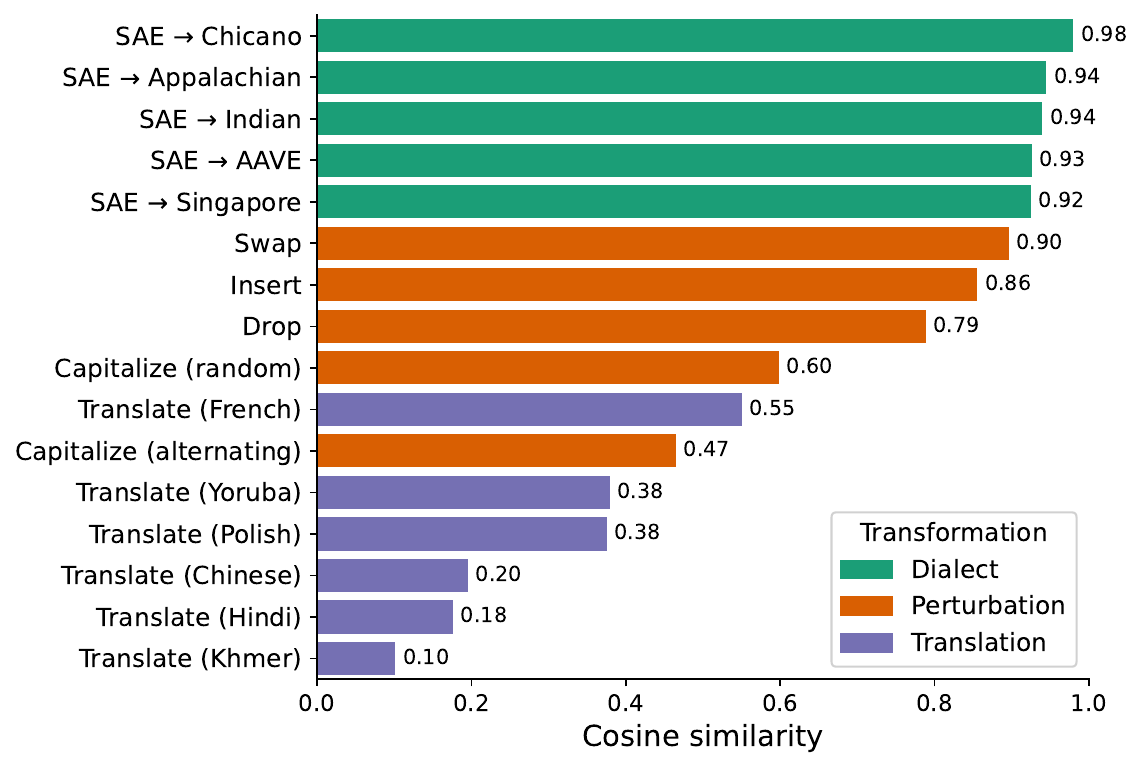}
\caption{\textbf{Final-layer hidden-state similarity.} The final-layer hidden-state similarity to that of SAE is pooled across \textsc{MultiVALUE} and \textsc{ParallelAAVE}.}
\label{figure:hidden_state_final_layer_ranking}
\end{figure}

Figure~\ref{figure:hidden_state_final_layer_ranking} visualizes the final-layer hidden-state similarity to SAE for each text transformation.

\section{Language Model Usage}
%%%%%%%%%%%%%
% LLM USAGE %
%%%%%%%%%%%%%

We used LMs for assistance with drafting and proofreading, as well as for code completions. They served as general-purpose research tools and did not make substantive contributions to the ideation, methodology, or content of this work. The authors take full responsibility for all aspects of the work.

%%%%%%%%%%%%%%%%%%%%%%%%%%%%%%

\end{document}